\documentclass[twoside,11pt]{article}
\usepackage[abbrvbib,nohyperref]{jmlr2e}

\usepackage{hyperref}
\hypersetup{
    colorlinks=true,
    linkcolor=blue,
    citecolor=blue,
    filecolor=magenta,
    urlcolor=green,
}
\usepackage{amsmath}
\usepackage{subcaption}
\usepackage{graphicx}
\usepackage{enumitem}
\usepackage{booktabs}
\usepackage{multirow}
\usepackage{makecell}
\usepackage{algorithm}
\usepackage{algorithmic}

\usepackage{tikz}
\usetikzlibrary{positioning, calc}

\def\f{Fr\'echet }
\DeclareMathOperator*{\argmin}{arg\,min}

\newtheorem{assumption}{Assumption}

\usepackage{lastpage}
\jmlrheading{27}{2026}{1-\pageref{LastPage}}{9/25; Revised 2/26}{3/26}{25-2364}{Yidong Zhou, Su I Iao and Hans-Georg M\"uller}

\ShortHeadings{End-to-End Metric Regression}{Zhou, Iao and M\"uller}
\firstpageno{1}

\begin{document}

\title{End-to-End Deep Learning for Predicting Metric Space-Valued Outputs}

\author{\name Yidong Zhou\thanks{The first two authors contributed equally to this work.} \email ydzhou@ucdavis.edu \\
       \addr Department of Statistics\\
       University of California, Davis\\
       Davis, CA 95616, USA
       \AND
       \name Su I Iao\footnotemark[1] \email siao@ucdavis.edu \\
       \addr Department of Statistics\\
       University of California, Davis\\
       Davis, CA 95616, USA
       \AND
       \name Hans-Georg M\"uller \email hgmueller@ucdavis.edu \\
       \addr Department of Statistics\\
       University of California, Davis\\
       Davis, CA 95616, USA}

\editor{Aryeh Kontorovich}

\maketitle

\begin{abstract}%   <- trailing '%' for backward compatibility of .sty file
Many modern applications involve predicting structured, non-Euclidean outputs such as probability distributions, networks, and symmetric positive-definite matrices. These outputs are naturally modeled as elements of general metric spaces, where classical regression techniques that rely on vector space structure no longer apply. We introduce E2M (End-to-End Metric regression), a deep learning framework for predicting metric space-valued outputs. E2M performs prediction via weighted \f means over training outputs, where the weights are learned by a neural network conditioned on the input. This construction provides a principled mechanism for geometry-aware prediction that avoids surrogate embeddings and restrictive parametric assumptions, while fully preserving the intrinsic geometry of the output space. We establish theoretical guarantees, including a universal approximation theorem that characterizes the expressive capacity of the model and a convergence analysis of the entropy-regularized training objective. Through extensive simulations involving probability distributions, networks, and symmetric positive-definite matrices, we show that E2M consistently achieves state-of-the-art performance, with its advantages becoming more pronounced at larger sample sizes. Applications to human mortality distributions and New York City taxi networks further demonstrate the flexibility and practical utility of this framework.
\end{abstract}


\section{Introduction}
The rapid growth of complex, structured data across science and engineering increasingly challenges traditional learning paradigms and demands fundamentally new modeling tools. In fields such as neuroscience \citep{dryd:09}, social science \citep{li:23}, and genomics \citep{kapl:20}, observations are increasingly recorded as non-Euclidean entities, which have been referred to as \emph{random objects} \citep{mull:16:7}.  Examples include functional data \citep{mull:16:3}, networks \citep{mull:22:11}, trees \citep{nye:17}, probability distributions \citep{pete:22}, and data residing in manifolds such as symmetric positive-definite (SPD) matrices \citep{huan:17}. These data types can be modeled as elements of general metric spaces, where the lack of algebraic operations such as addition, subtraction, or scalar multiplication renders standard statistical and machine learning techniques inapplicable.

\begin{figure}[tb]
    \centering
    \begin{subfigure}{0.45\textwidth}
        \centering
        \includegraphics[width=1\linewidth]{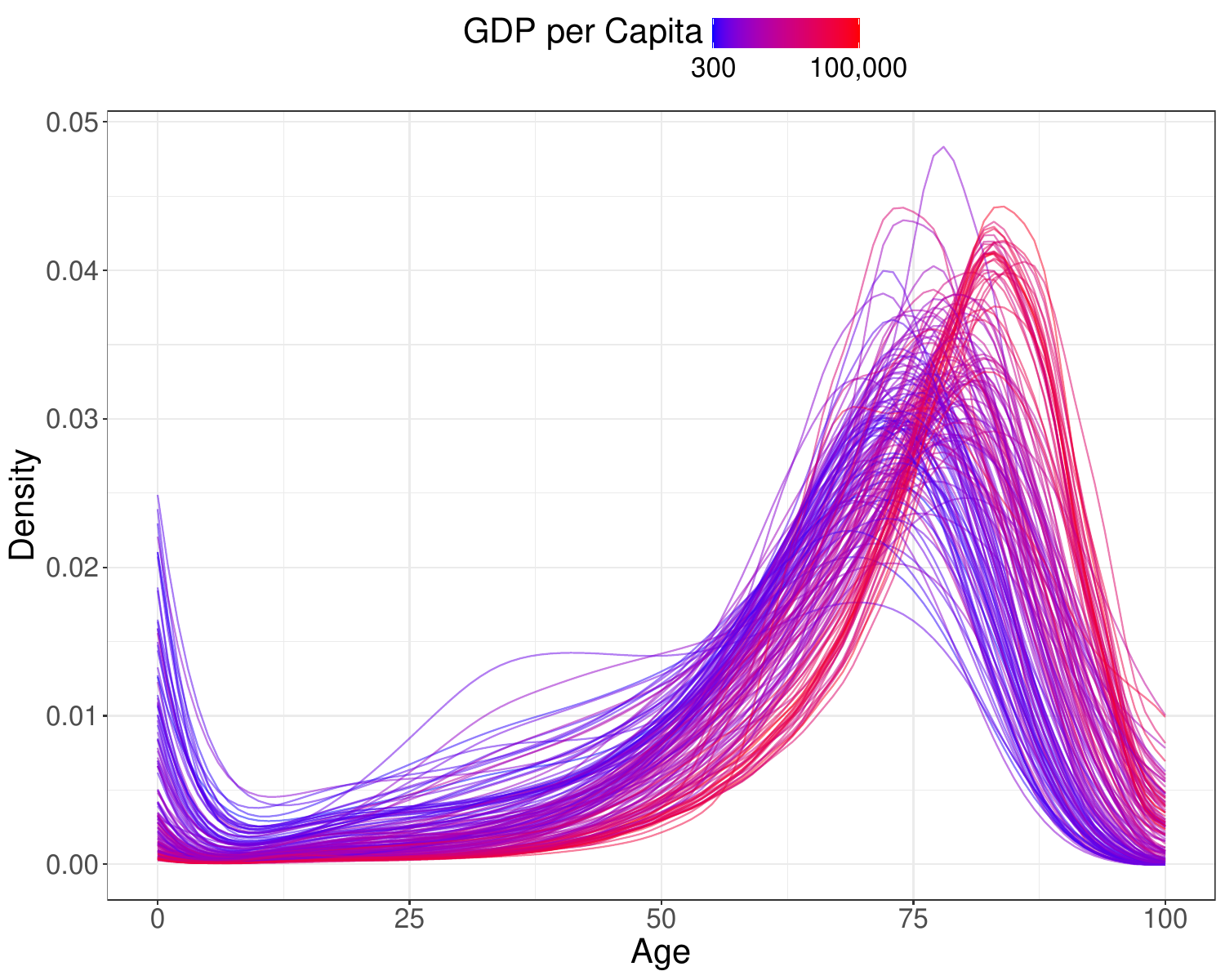}
    \end{subfigure}\hfill
    \begin{subfigure}{0.45\textwidth}
        \centering
        \includegraphics[width=1\linewidth]{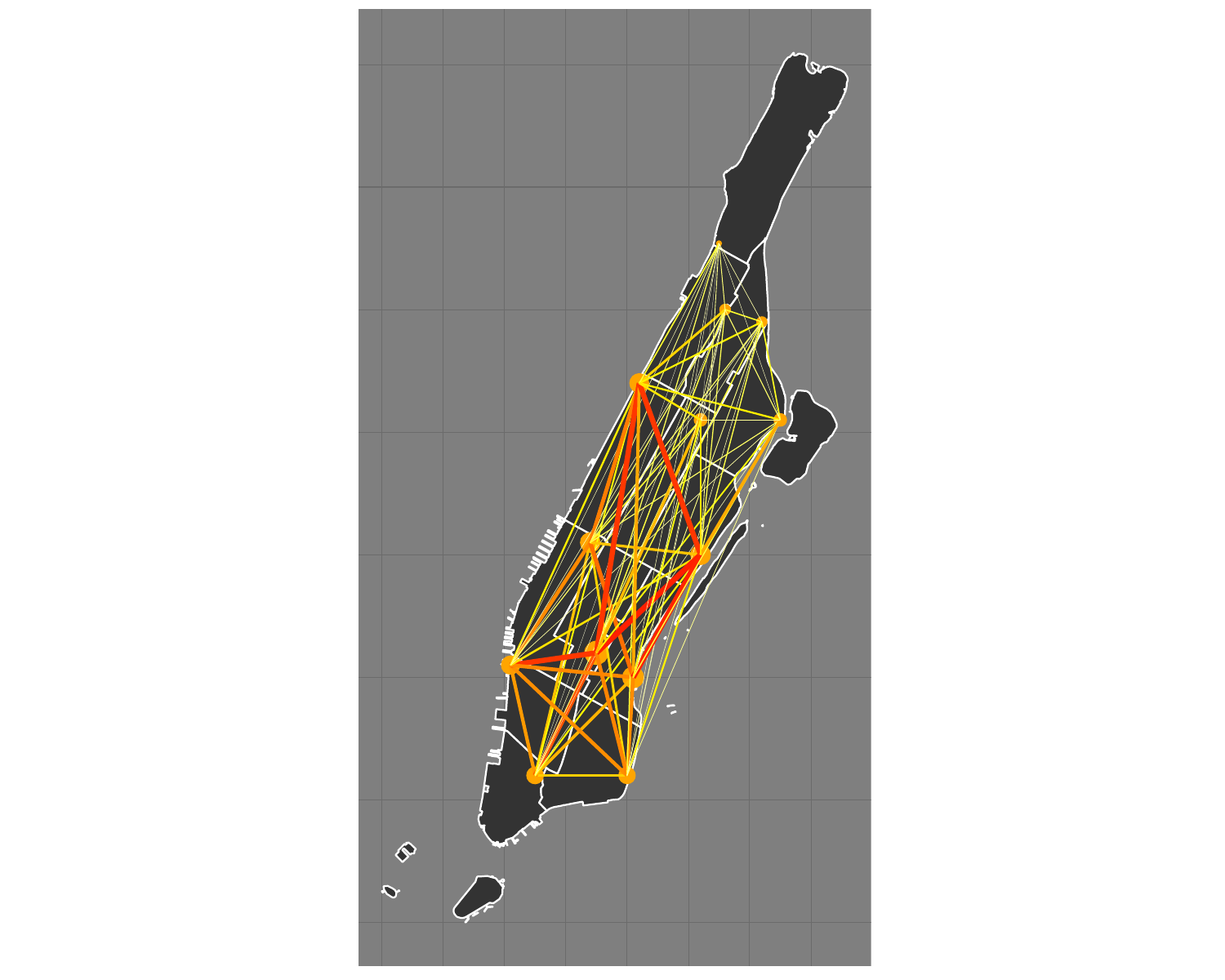}
    \end{subfigure}
    \caption{Motivating examples of non-Euclidean outputs. Left: age-at-death densities for 162 countries in 2015, colored by GDP per capita. Right: daily transportation network in Manhattan on January 1, 2018.}
    \label{fig:motivation}
\end{figure}

To motivate the problem, Figure~\ref{fig:motivation} illustrates two representative examples. The left panel shows age-at-death densities for 162 countries in 2015, with curves colored by GDP per capita. Here, the regression problem of interest is to model how the age-at-death distribution varies with demographic, economic, and environmental indicators, enabling comparisons across populations with different socioeconomic conditions. The right panel displays a daily traffic network in Manhattan on January 1, 2018, constructed from New York City yellow taxi trip data. In this setting, the goal is to predict the structure of daily transportation networks from predictors such as weather conditions, calendar effects, and aggregated trip statistics. Both examples highlight outputs that are complex objects such as probability distributions and networks, rather than vectors; such objects are inherently non-Euclidean.

Inspired by these applications, we study supervised learning when the input is a vector $X \in \mathbb{R}^p$ and the output $Y$ resides in a general metric space $(\Omega, d)$. While the proposed methodology applies to general metric spaces, the theoretical results developed in Section~\ref{sec:the} require additional assumptions for the space $(\Omega, d)$, including boundedness and for some results that it is a Hadamard space. These assumptions are used to establish continuity, Lipschitz properties, and optimization guarantees. Learning in this setting is fundamentally more challenging than for vector-valued outputs due to the lack of vector space structure in $\Omega$. Existing solutions often rely on Euclidean embeddings \citep{fara:14,zhan:21:1,iao:25} or restrictive model assumptions \citep{hein:09,mull:19:6,han:23}, which either distort the geometry of the output space or lack robustness in real-world scenarios. These limitations highlight the need for principled intrinsic methods that operate directly on non-Euclidean data, preserving the structural characteristics of the outputs while avoiding oversimplifying assumptions.

In this paper, we propose \textbf{E2M} (End-to-End Metric regression), a deep learning framework for regression with outputs in general metric spaces. E2M revisits the classical idea of regression as a weighted average over training outputs, adapting it to metric spaces using the \emph{weighted \f mean} \citep{frec:48}. The weighted \f mean minimizes a weighted sum of squared distances and remains well-defined without requiring algebraic operations within the output space. In E2M, a neural network conditioned on the input generates weights that encode the relevance of each training output in the weighted \f mean computation. This approach preserves the intrinsic geometry of the output space while enabling flexible data-driven prediction.

Our main contributions are as follows:
\begin{itemize}
    \item We propose E2M, the first end-to-end deep learning framework for supervised learning with metric space-valued outputs. E2M uses a neural network to learn sample-specific weights and constructs predictions via a weighted \f mean. The key idea is to represent the regression function as a weighted average over training outputs. This formulation circumvents the lack of vector space operations in the output space and avoids reliance on surrogate embeddings. The model preserves output geometry and incorporates entropy regularization to control the sparsity and adaptivity of the weights.
    \item We establish theoretical guarantees for E2M, including a universal approximation theorem that characterizes the expressive power of the model and convergence results for the entropy-regularized training objective. Unlike standard regression settings, the analysis is complicated by the lack of linear structure in the output space. To address this challenge, we employ tools from metric geometry, including properties of \f means and variance inequalities in Hadamard spaces \citep{stur:03}, to prove Lipschitz continuity and derive optimization guarantees.
    \item We demonstrate the effectiveness of E2M through extensive simulations and real-world applications involving probability distributions, networks, and SPD matrices. Across various scenarios, E2M consistently outperforms existing regression methods for non-Euclidean targets, demonstrating its flexibility, accuracy, and geometric robustness.
\end{itemize}

\section{Related Work}
Our work intersects several research areas, including neural networks for adaptive weighting, geometric deep learning and regression for metric space-valued outputs. We briefly review these areas and highlight how E2M conveys major advances over existing approaches.

\paragraph{Neural networks for weighted predictions.} Neural networks often produce adaptive weights as part of their prediction pipeline. Attention mechanisms, introduced in Transformers \citep{vasw:17}, use neural networks to compute softmax-normalized weights for aggregating context-aware representations. Similarly, geographically weighted neural networks \citep{hage:22} assign spatially varying weights to combine neural outputs based on input location. However, in these approaches, weights are used to aggregate intermediate features, and predictions are typically constrained to Euclidean spaces. In contrast, E2M directly generates predictions in general metric spaces using neural network-derived weights that determine a weighted \f mean of training outputs.

\paragraph{Geometric deep learning.} This line of work adapts deep learning to structured input domains such as graphs, manifolds, and sets \citep{bron:17}. Methods like graph neural networks \citep{kipf:17} and manifold-based networks \citep{chak:20} effectively handle complex input geometries for tasks like graph node classification or shape analysis. However, these techniques generally assume outputs reside in Euclidean spaces, with geometric structure considered only in input space. Our work addresses the complementary and much less explored problem of predicting outputs that reside in general metric spaces, which is relevant for many real-world applications, thus broadening the scope of geometry-aware modeling.

\paragraph{Regression models for metric space-valued outputs.} A growing body of work extends learning and regression to cover metric space-valued outputs. Early methods involved embedding metric spaces into Euclidean frameworks \citep{fara:14} or applying kernel-based techniques \citep{hein:09}. More recently, \f regression generalized linear and local linear regression to metric space-valued outputs \citep{mull:19:6}. Subsequent extensions of this framework include methods for sufficient dimension reduction \citep{ying:22,zhan:21:1}, single-index models \citep{mull:23:3,ghos:23}, principal component regression \citep{han:23}, and adaptations to tree-based approaches \citep{capi:19,qiu:24,bult:24,zhou:25}. Related learning-theoretic work studies Bayes consistency in general metric spaces, including universal Bayes consistency for nearest-neighbor--type classification rules under metric inputs \citep{hann:21} and Bayes-consistent prediction under general metric losses using data-supported, Fr\'echet-type aggregation rules \citep{cohe:22}. A recent method \citep{iao:25} incorporates neural networks into \f regression in a three-step approach, where neural networks are used to model the relationship between the Euclidean input and the low-dimensional manifold representation of the metric space-valued output. While effective, this approach relies on a low-dimensional manifold assumption, which may not hold in practice. In contrast, E2M provides a fully end-to-end learning framework, where both feature extraction and predictive weighting are jointly optimized via backpropagation, without restrictive assumptions on the intrinsic dimensionality of the output space. 

\section{Preliminaries}
Let $(\Omega, d)$ be a metric space and consider a random object $Y$ taking values in $\Omega$. The classical concept of expectation in Euclidean space extends naturally to this setting via the \f mean \citep{frec:48}, defined by
\[E_\oplus[Y]=\argmin_{y\in\Omega}E[d^2(y, Y)],\]
where the existence and uniqueness of the minimizer depend on the geometry of the underlying metric space and are guaranteed in Hadamard spaces \citep{stur:03}; see Definition~\ref{def:hs}.

To illustrate the scope of E2M, we present several representative examples of metric spaces that arise frequently in modern applications. These examples are used throughout our simulations and empirical analyses to demonstrate the generality and flexibility of the proposed framework.

\begin{example}[One-dimensional probability distributions]
\label{exm:mea}
Consider the Wasserstein space $(\mathcal{W}, d_{\mathcal{W}})$ \citep{pana:20} of one-dimensional probability distributions with finite second moments, equipped with the 2-Wasserstein metric $d_{\mathcal{W}}$. The 2-Wasserstein metric between two distributions $\mu_1$ and $\mu_2$ is
\[
d_{\mathcal{W}}^2(\mu_1, \mu_2)=\int_0^1\{F_{\mu_1}^{-1}(p)-F_{\mu_2}^{-1}(p)\}^2\,dp,
\]
where $F_{\mu_1}^{-1}$ and $F_{\mu_2}^{-1}$ are the quantile functions of $\mu_1$ and $\mu_2$, respectively. The Wasserstein space has attracted considerable attention across statistics and data science, as the Wasserstein metric provides a meaningful similarity measure between probability distributions that reflects the geometry of the underlying sample space \citep{vill:03}. It has been widely used in modern applications, including Wasserstein generative adversarial networks \citep{arjo:17} and Wasserstein autoencoders \citep{tols:18}, where it enables more stable training and better capture of distributional structures compared to classical divergence-based approaches.
\end{example}

\begin{example}[Networks]
\label{exm:net}
Consider the space of simple, undirected, weighted networks with a fixed number of nodes and bounded edge weights. Each network can be represented uniquely by its graph Laplacian. The space of graph Laplacians equipped with the Frobenius metric can thus be used to characterize the space of networks \citep{kola:14,seve:19,mull:22:11}. Graph Laplacians have been widely used in both spectral methods and modern approaches to network representation and learning. For example, the graph convolutional network \citep{kipf:17} builds convolutional operations directly from the graph Laplacian, highlighting its role in capturing the intrinsic geometry of network data.
\end{example}

\begin{example}[Symmetric positive-definite matrices]
\label{exm:mat}
Consider the space of $l \times l$ symmetric positive-definite (SPD) matrices, denoted $\mathrm{Sym}^+_l$, with important examples including covariance and correlation matrices. Various metrics endow $\mathrm{Sym}^+_l$ with rich geometric structure, including the Frobenius metric, the affine-invariant metric \citep{penn:06}, the power metric \citep{dryd:09}, the Log-Cholesky metric \citep{lin:19:1}, and the Bures--Wasserstein (BW) metric \citep{bhat:19}. The non-Euclidean geometry of $\mathrm{Sym}^+_l$ plays a crucial role in machine learning and signal processing and recent methods have incorporated its Riemannian manifold structure directly into learning frameworks. A prominent example is SPDNet \citep{huan:17}, a deep neural network where intermediate layers preserve the manifold structure through bilinear and eigenvalue operations and the final logarithm mapping layer projects SPD matrices into Euclidean features for downstream tasks such as classification.
\end{example}

In the presence of predictors $X \in \mathbb{R}^p$, one can consider conditional \f means \citep{mull:19:6}:
\[E_\oplus(Y|X = x) = \argmin_{y \in \Omega}E[d^2(y, Y)|X = x],\]
where the expectation is taken with respect to the conditional distribution of $Y$ given $X$. This definition generalizes the classical conditional expectation, which is recovered when $\Omega = \mathbb{R}$ and $d$ is the Euclidean distance. A more detailed discussion of the relationship between classical and \f means is provided in Appendix~\ref{app:mean}. As in classical regression, the conditional \f mean $m(x)=E_\oplus(Y|X=x)$ serves as the target for regression with metric space-valued outputs.

A wide range of classical regression methods estimate the regression function by expressing predictions as weighted averages of the observed outputs. Notable examples include Nadaraya--Watson kernel regression \citep{nada:64, wats:64}, $k$-nearest neighbors regression \citep{altm:92}, inverse distance weighting \citep{shep:68}, linear regression, and local linear regression \citep{fan:96}. Given training samples $\{(X_i, Y_i)\}_{i=1}^n$ and an input $x$, these methods assign a weight $w_i(x)$ to each training pair $(X_i, Y_i)$ based on the proximity between $x$ and $X_i$. Specifically, when $Y_i \in \mathbb{R}$, the prediction can be interpreted as a weighted average:
\[\hat{m}(x)=\sum_{i=1}^n w_i(x) Y_i=\argmin_{y \in \mathbb{R}}\sum_{i=1}^n w_i(x)(Y_i - y)^2.\]
This variational form reveals that such methods estimate $m(x)$ by minimizing a weighted squared loss, where the weights encode relevance via predictor similarity. Further details on how classical linear and local linear regression also admit such characterizations are provided in Appendix~\ref{app:reg}.

This weighted formulation generalizes naturally to metric space-valued outputs by replacing the squared Euclidean loss $(Y_i - y)^2$ with a squared distance induced by the metric $d$. The resulting prediction takes the form of a weighted \f mean,
\[\hat{m}(x) = \argmin_{y \in \Omega} \sum_{i=1}^n w_i(x)\, d^2(y, Y_i).\]
Weighted \f means and related barycenter constructions have been extensively studied in the literature, including weighted Wasserstein barycenters \citep{ague:11,lego:17} and \f regression \citep{mull:19:6}. This generalization preserves the intuition of proximity-weighted averaging while accommodating outputs in structured or nonlinear spaces. A key advantage is that the learned weights are directly interpretable: each $w_i(x)$ quantifies the contribution of training sample $Y_i$ to the prediction at input $x$. Concentrated weights highlight locally influential samples, whereas more uniform weights reflect global smoothing. In this way, the method provides insight into sample relevance that is difficult to obtain from models that compress training information into fixed parameters.

\section{Methodology}

\subsection{E2M}
Rather than specifying the weight function $w_i(x)$ through fixed kernels or distance rules, we propose to learn it directly from data. To this end, we introduce E2M, a regression framework for metric space-valued outputs, where a neural network is trained end-to-end to generate adaptive, task-specific weights that minimize prediction error in the target metric space. The architecture of the proposed model is illustrated in Figure~\ref{fig:e2m}.

\begin{figure}[tb]
\centering
\begin{tikzpicture}[->, thick, node distance=1.7cm and 1.5cm]

% Input node
\node[circle, draw=black, fill=green!20, minimum size=1cm] (X) {$x$};

% First layer (outputs of g)
\node[circle, draw=black, fill=blue!20, minimum size=0.8cm, right=2cm of X, yshift=1.3cm] (g1) {$z_1$};
\node[circle, draw=black, fill=blue!20, minimum size=0.8cm, right=2cm of X] (g2) {$z_i$};
\node[circle, draw=black, fill=blue!20, minimum size=0.8cm, right=2cm of X, yshift=-1.3cm] (g3) {$z_n$};

% softmax function box
\node[rectangle, draw=black, fill=orange!20, minimum width=0.8cm, minimum height=3.5cm, right=1cm of g2] (softmax) {Softmax};

% Second layer (after softmax)
\node[circle, draw=black, fill=yellow!30, minimum size=0.8cm, right=1cm of softmax, yshift=1.3cm] (s1) {$w_1$};
\node[circle, draw=black, fill=yellow!30, minimum size=0.8cm, right=1cm of softmax] (s2) {$w_i$};
\node[circle, draw=black, fill=yellow!30, minimum size=0.8cm, right=1cm of softmax, yshift=-1.3cm] (s3) {$w_n$};

% Final output node
\node[circle, draw=black, fill=red!20, minimum size=1cm, right=3.2cm of s2] (Yhat) {$m_\theta(x)$};

% Arrows from X to g outputs
\draw (X) -- (g1);
\draw (X) -- (g2);
\draw (X) -- (g3);

% % Arrows from g outputs to softmax outputs
% \draw (g1) -- (s1);
% \draw (g2) -- (s2);
% \draw (g3) -- (s3);
% Arrows from g outputs to softmax box
\draw (g1) -- (softmax.west |- g1);
\draw (g2) -- (softmax.west);
\draw (g3) -- (softmax.west |- g3);

% Arrows from softmax box to w outputs
\draw (softmax.east |- g1) -- (s1);
\draw (softmax.east) -- (s2);
\draw (softmax.east |- g3) -- (s3);

% Arrows from softmax outputs to final output
\draw (s1) -- (Yhat);
\draw (s2) -- (Yhat);
\draw (s3) -- (Yhat);

% Dots between g1 and g3
\node at ($(g1)!0.42!(g2)$) {$\vdots$};
\node at ($(g2)!0.42!(g3)$) {$\vdots$};

% Dots between s1 and s3
\node at ($(s1)!0.42!(s2)$) {$\vdots$};
\node at ($(s2)!0.42!(s3)$) {$\vdots$};

% Mappings on top
\node[align=center, above=1.3cm of X] (labelX) {$\mathbb{R}^p$};
% \node[align=center, above=1.4cm of g2] (labelG) {$\mathbb{R}^n$};
\node[align=center, above=1.4cm of s2] (labelDelta) {$\Delta^{n-1}$};
\node[align=center, above=1.16cm of Yhat] (labelOmega) {$\Omega$};

% Arrows for mappings (above, between spaces)
% \draw[->] (labelX) -- node[above, yshift=0cm] {\footnotesize neural networks: $g_\theta$} (labelG);
% \draw[->] (labelG) -- node[above, yshift=0cm] {\footnotesize softmax: $w$} (labelDelta);
\draw[->] (labelX) -- node[above, yshift=0cm] {\footnotesize neural network: $w_\theta$} (labelDelta);
\draw[->] (labelDelta) -- node[above, yshift=0cm] {\footnotesize weighted \f mean: $\mu$}
(labelOmega);
% \draw[->, bend left=40] (labelG.north east) to node[above, yshift=0.1cm] {\footnotesize $h$} (labelOmega.north west);
\end{tikzpicture}
\caption{Schematic diagram for E2M. Here  $m_\theta= \mu \circ w_\theta$, where $w_\theta$ is a neural network parameterized by $\theta$ and $\mu$ denotes the weighted \f mean. The input $x \in \mathbb{R}^p$ is mapped to $\mathbb{R}^n$, passed through softmax to $\Delta^{n-1}$ via neural network $w_\theta$ and then mapped via the weighted \f mean $\mu$ to the metric space $\Omega$.}
\label{fig:e2m}
\end{figure}

The model is $m_\theta = \mu \circ w_\theta$, where the first component, $w_\theta: \mathbb{R}^p \mapsto \Delta^{n-1}$, is a fully connected neural network with multiple hidden layers using rectified linear unit (ReLU) activations, followed by a softmax output layer. Here $\theta$ represents the network parameters, and $\Delta^{n-1} = \left\{ w \in \mathbb{R}^n : w_i \geq 0 \text{ for all } i,\ \sum_{i=1}^n w_i = 1 \right\}$ is the $(n-1)$-simplex. Given an input $x$, the network outputs a weight vector $w_\theta(x) \in \Delta^{n-1}$ that assigns relevance scores to each of the training samples.
The second component, $\mu: \Delta^{n-1} \mapsto \Omega$, is a weighted \f mean,
\[\mu(w) = \argmin_{y \in \Omega} \sum_{i=1}^n w_i d^2(y, Y_i),\]
mapping the learned weights to a prediction in the metric space $\Omega$, and  $w_i$ denotes the $i$th coordinate of the softmax output.

By learning $w_\theta$ end-to-end together with the weighting and prediction mechanism, E2M bypasses the need for hand-crafted feature engineering or local distance computations. To learn the optimal network parameters, we minimize the empirical loss
\[\frac{1}{n} \sum_{i=1}^n d^2(m_\theta(X_i), Y_i),\]
which measures the discrepancy between the predicted outputs and the observed outputs in the metric space $\Omega$. The learned network implicitly captures statistical dependencies between the input and the output. This feedback mechanism allows the model to go beyond proximity-based heuristics and discover weighting schemes that reflect predictive relevance. In particular, E2M can assign low weights to samples that are geometrically close but uninformative or noisy with respect to the target.

\subsection{Regularization}
To enhance the stability and generalization of the learned model, E2M incorporates regularization techniques that directly act on the learned weight distribution. We primarily adopt entropy regularization, which provides a principled and interpretable mechanism to control the sharpness and dispersion of the weights.
The empirical loss with entropy regularization is defined as
\[\mathcal{L}_{n}(\theta) = \frac{1}{n}\sum_{i=1}^n d^2( m_\theta(X_i), Y_i ) + \lambda \frac{1}{n}\sum_{i=1}^n H(w_\theta(X_i)),\]
where $H(w) = -\sum_{i=1}^n w_i \log(w_i + \delta)$ denotes the entropy of the weight vector, $\lambda \in \mathbb{R}$ is a hyperparameter controlling the strength of regularization, and $\delta$ is a small positive constant for numerical stability (set to $10^{-10}$ in the implementation).

Entropy regularization offers a simple yet effective way to control the behavior of the model through the concentration of weights. Positive values of $\lambda$ encourage sharper, low-entropy distributions that prioritize a small subset of influential training samples---akin to local regression. In contrast, negative values of $\lambda$ promote higher-entropy, more uniform weight distributions, leading to a global smoothing effect. This flexibility allows E2M to adapt to varying data structures and modeling needs. In practice, the regularization parameter $\lambda$ is chosen through cross-validated grid search; see Appendix~\ref{app:hyper} for details. To further assess robustness, we conducted a sensitivity analysis, reported in Appendix~\ref{app:entropy}, which demonstrates that performance is stable over a range of values. The full training procedure is outlined in Algorithm~\ref{alg:e2m}.

\begin{algorithm}[t]
\caption{E2M Training Procedure}
\label{alg:e2m}
\begin{algorithmic}[1]
\REQUIRE Training samples $\{(X_i, Y_i)\}_{i=1}^n$, entropy regularization parameter $\lambda$
\ENSURE Trained neural network $w_\theta$
\STATE Initialize parameters $\theta$ of the neural network $w_\theta$
\STATE Optimize $\theta$ using the Adam optimizer with entropy-regularized empirical loss:
\[\mathcal{L}_{n}(\theta) = \frac{1}{n}\sum_{i=1}^n d^2( m_\theta(X_i), Y_i ) + \lambda \frac{1}{n}\sum_{i=1}^n H(w_\theta(X_i)),\]
where $m_\theta=\mu\circ w_\theta$ and $\mu(w)=\argmin_{y\in\Omega} \sum_{i=1}^nw_id^2(y, Y_i)$
\end{algorithmic}
\end{algorithm}

\section{Theory}\label{sec:the}
We analyze the theoretical properties of the E2M framework using tools from metric geometry and optimization. We begin by proving a universal approximation theorem, which establishes the expressive capacity of the model. Next, we show that the weighted \f mean is Lipschitz continuous with respect to the weight vector, using a variance inequality from metric geometry \citep{stur:03}. This Lipschitz property is instrumental in analyzing the convergence of the training algorithm. Proofs of all lemmas and theorems are provided in Appendix~\ref{app:proof}.

We consider the target mapping $m = \mu \circ w$, where $w: \mathbb{R}^p \mapsto \Delta^{n-1}$ is a continuous function and $\mu: \Delta^{n-1} \mapsto \Omega$ is a deterministic map corresponding to the weighted \f mean. The learning task is to approximate $w$ using a neural network $w_\theta$, so that $m_\theta = \mu \circ w_\theta$ approximates $m$.

\begin{assumption}\label{ass:uniq}
    For every $w\in\Delta^{n-1}$, the weighted \f mean $\mu(w)$ exists and is unique.
\end{assumption}

This assumption guarantees that the map $\mu$ is well-defined and holds for the spaces described in Examples~\ref{exm:mea} and \ref{exm:net}. For Example~\ref{exm:mat} it will depend on the specific metric that one adopts whether the assumption holds or not. It holds when equipping the space with any of the metrics mentioned except for the BW metric and even then it may be satisfied, depending on the specifics of the sample. For simulated data we found it to be still satisfied for most realizations. Under this condition, we apply the Berge Maximum Theorem~\citep[Theorem 17.31]{alip:06} to establish the continuity of the weighted \f mean map $\mu$.

\begin{lemma}\label{lem:frechet-continuity}
Suppose Assumption~\ref{ass:uniq} holds and $(\Omega, d)$ is compact. Then the weighted \f mean map $\mu: \Delta^{n-1} \mapsto \Omega$ is continuous on $\Delta^{n-1}$.
\end{lemma}

This continuity enables the following universal approximation result.

\begin{theorem}\label{thm:universal-approximation}
Suppose Assumption~\ref{ass:uniq} holds and $(\Omega, d)$ is compact. For any $\epsilon > 0$, there exists a neural network $w_{\theta^*}$ such that the function $m_{\theta^*} = \mu\circ w_{\theta^*}$ satisfies
\[\sup_{\|x\| \leq 1} d(m_{\theta^*}(x), m(x)) < \epsilon.\]
If $X$ is stochastically bounded, then for any $\delta > 0$ there exists a neural network $w_{\theta^*}$ such that
\[P\big(d(m_{\theta^*}(X), m(X)) < \epsilon\big) > 1 - \delta.\]
\end{theorem}

Next, we analyze the convergence behavior of the proposed architecture within the geometric framework of Hadamard spaces, which provide a natural setting for our analysis.

\begin{definition}[Hadamard space]\label{def:hs}
A metric space $(\Omega, d)$ is called a \textit{Hadamard space} if it is complete and if for each pair of points $\omega_1, \omega_2 \in \Omega$, there exists a point $\alpha \in \Omega$ satisfying:
\[d^2(\beta,\alpha) \leq \frac{1}{2} d^2(\beta,\omega_1) + \frac{1}{2} d^2(\beta,\omega_2) - \frac{1}{4}d^2(\omega_1,\omega_2), \quad \text{for all } \beta \in \Omega.\]
\end{definition}

Hadamard spaces, also known as globally non-positively curved spaces~\citep{stur:03}, are uniquely geodesic and admit well-behaved notions of distance and convexity. In particular, the weighted \f mean $\mu(w)$ is guaranteed to exist and be unique for any weight vector $w \in \Delta^{n-1}$~\citep{baca:14}. Common examples of Hadamard spaces include Euclidean and Hilbert spaces, hyperbolic spaces and various other spaces frequently encountered in applications, including those discussed in Examples~\ref{exm:mea} and \ref{exm:net}. In Example~\ref{exm:mat}, the space of SPD matrices equipped with any of the metrics mentioned there except for the BW metric is Hadamard. These spaces have been widely studied in optimization~\citep{baca:14:2}, regression~\citep{mull:21:5} and geometric deep learning~\citep{gane:18}.
\begin{remark}
    The theoretical analysis assumes that the output space $(\Omega,d)$ is a Hadamard space. This condition is mainly required for theory, as it guarantees convexity and variance inequalities that yield Lipschitz continuity of the weighted \f mean map and enable convergence analysis. Several experimental settings in this paper, such as the Wasserstein space for univariate distributions, SPD matrices with power metrics and networks with the Frobenius metric satisfy this property. To further examine the scope of the method, an additional simulation in Section~\ref{sec:sim} with SPD matrices under the BW metric, a positively curved space that is not Hadamard \citep{than:23}, shows that E2M continues to perform effectively. Thus, while the Hadamard assumption is important for establishing theoretical guarantees, the method remains practically applicable in a broader class of metric spaces. From a theoretical perspective, extending the analysis to metric spaces where weighted \f means may fail to be unique, or where the \f functional is not geodesically convex, remains an important open problem. Developing guarantees for optimization and statistical behavior of E2M under such conditions is a promising direction for future research.
\end{remark}

\begin{algorithm}[t]
\caption{Adam}
\label{alg:adam}
\begin{algorithmic}[1]
\REQUIRE Initial parameter $\theta_0$, learning rate $\eta$, decay parameters $\beta_1, \beta_2\in[0, 1]$, $\epsilon > 0$
\STATE Set $m_0 = 0$, $v_0 = 0$
\FOR{$k = 1$ to $T$}
    \STATE Draw a minibatch size of $b$: $\{(X_j, Y_j)\}_{j=1}^b$
    \STATE Compute $g_k = \frac{1}{b}\sum_{j=1}^b\nabla \ell(\theta_k; (X_j, Y_j))$
    \STATE $m_k = \beta_1 m_{k-1} + (1 - \beta_1)g_k$
    \STATE $v_k = \beta_2v_{k-1} + (1 - \beta_2)g_k^2$
    \STATE $\hat{m}_k=m_k/(1-\beta_1^k)$
    \STATE $\hat{v}_k=v_k/(1-\beta_2^k)$
    \STATE $\theta_k = \theta_{k-1} - \eta \hat{m}_k / (\sqrt{\hat{v}_k} + \epsilon)$
\ENDFOR
\end{algorithmic}
\end{algorithm}

We train the neural network $w_\theta$ using the Adam optimization algorithm \citep{king:15} to minimize the entropy-regularized empirical loss. The procedure is summarized in Algorithm~\ref{alg:adam}. Let $\ell(\theta; (X, Y)) = d^2(m_\theta(X), Y) + \lambda H(w_\theta(X))$ denote the per-sample loss with entropy regularization and let $\mathcal{L}(\theta) = E[\ell(\theta; (X, Y))]$ be the corresponding population loss. While $\mathcal{L}$ is convex in certain special cases (e.g., Examples~\ref{exm:mea}--\ref{exm:mat}), it is generally non-convex due to the nested minimization in the weighted \f mean $\mu$. To address this, we establish a Lipschitz bound for the weighted \f mean map using the variance inequality~\citep{stur:03}.

\begin{lemma}\label{lem:frechet-lipschitz}
If $(\Omega, d)$ is a bounded Hadamard space, then the weighted \f mean map $\mu: \Delta^{n-1} \to \Omega$ is Lipschitz continuous on $\Delta^{n-1}$. Specifically, for any $w_1, w_2 \in \Delta^{n-1}$, we have
\[
d(\mu(w_1), \mu(w_2)) \leq D\sqrt{n} \, \|w_1 - w_2\|_2,
\]
where $D = \sup_{u,v \in \Omega}d(u,v)$ is the diameter of $\Omega$.
\end{lemma}

Lemma~\ref{lem:frechet-lipschitz} implies the Lipschitz continuity of the loss function $\ell(\theta; (X, Y))$, a property essential for convergence analysis, and is broadly useful for studying the stability of weighted \f means with respect to weights.

\begin{remark}[Scalability via anchors]\label{rem:anchor}
    The Lipschitz constant in Lemma~\ref{lem:frechet-lipschitz} contains a $\sqrt{n}$ factor. While this may appear undesirable at first glance, note that $n$ here refers to the number of anchor points used in computing the weighted \f mean, not necessarily the overall sample size. For simplicity, the implementation uses all training outputs $\{Y_i\}_{i=1}^n$ as anchors, hence the notation. However, our framework does not require this choice. In large-scale settings, one can subsample a fixed set of anchors independent of the dataset size, in which case the Lipschitz constant becomes independent of the total sample size. This anchor-based strategy provides a natural extension for scalability and Section~\ref{subsec:anchor} demonstrates through large-scale experiments that it preserves predictive accuracy while substantially improving computational efficiency.
\end{remark}

\begin{assumption}\label{ass:generalization}
We impose the following assumptions:
\begin{enumerate}[label=(\roman*)]
    \item \textbf{Lipschitz neural network:} The neural network $w_\theta: \mathbb{R}^p \mapsto \Delta^{n-1}$ is $L$-Lipschitz continuous with respect to its parameters $\theta$.
    \item \textbf{Smoothness:} The loss $\ell(\theta; (X, Y))$ is $\beta$-smooth with respect to $\theta$, i.e., its gradient $\nabla_\theta \ell(\theta; (X, Y))$ is $\beta$-Lipschitz continuous.
    \item \textbf{Bounded variance:} The variance of the stochastic gradient is bounded, i.e., for some $\sigma^2>0$ it holds that $E[\|\nabla_\theta \ell(\theta; (X, Y)) - \nabla \mathcal{L}(\theta)\|_2^2] \leq \sigma^2$.
\end{enumerate}
\end{assumption}

These assumptions are standard in non-convex optimization settings \citep{zahe:18, chen:19:2}. Assumption~\ref{ass:generalization}(i) is mild, as neural networks are Lipschitz continuous when weight matrices are bounded and the activation functions are Lipschitz, which holds for commonly used choices such as ReLU, Tanh or Sigmoid. In practice, techniques such as spectral normalization \citep{miya:18}, weight clipping \citep{arjo:17} and Lipschitz regularization \citep{gouk:21} are frequently employed to enforce Lipschitz continuity of $w_\theta$.

\begin{theorem}\label{thm:adam}
Suppose $(\Omega, d)$ is a bounded Hadamard space and Assumption~\ref{ass:generalization} holds. Let Algorithm~\ref{alg:adam} be run with mini-batch size $b$ and hyperparameters satisfying $\eta \leq \epsilon/(2\beta)$ and $1 - \beta_2 \leq \epsilon^2/(16G^2)$, where
$G = L\sqrt{n} \left(2D^2 + |\lambda| (|\log \delta| + 1)\right)$
is the Lipschitz constant of the loss $\ell(\theta; (X, Y))$ with respect to $\theta$. Then for an iterate $\theta_\tau$ chosen uniformly at random from $\{\theta_1, \dots, \theta_T\}$, we have
\[
E\left[\|\nabla \mathcal{L}(\theta_\tau)\|_2^2\right] = O\left(\frac{1}{T} + \frac{1}{b}\right).
\]
\end{theorem}
This result shows that the algorithm converges to a stationary point, with the $1/T$ term reflecting the effect of training duration and the $1/b$ term capturing the variance reduction from larger mini-batches.

\section{Numerical Experiments}
\label{sec:sim}
We evaluate the performance of E2M through comprehensive simulations involving three types of non-Euclidean outputs: probability distributions modeled in the Wasserstein space with the Wasserstein metric (Example \ref{exm:mea}), networks represented by graph Laplacians with the Frobenius metric (Example \ref{exm:net}) and SPD matrices equipped either with the power metric with exponent $1/2$ or with the BW metric (Example \ref{exm:mat}). Each setting is tested across sample sizes $n=500,1000,2000$, with $200$ Monte Carlo replications per scenario. We compare the performance of E2M with that of deep \f regression (DFR) \citep{iao:25}, global \f regression (GFR) \citep{mull:19:6}, sufficient dimension reduction (SDR) \citep{zhan:21:1}, single-index \f regression (IFR) \citep{mull:23:3} and random forest weighted local \f regression (RFWLFR) \citep{qiu:24}. Note that SDR and IFR could not be applied to SPD outputs with the power metric because no compatible implementations are available. Moreover, current implementations of DFR, GFR, SDR, IFR and RFWLFR do not handle SPD outputs under the BW metric; extending them would require nontrivial adaptations that are specific to both the space and the metric. To further assess robustness, we include additional experiments evaluating performance in small sample regimes (Appendix~\ref{app:small}) and in settings where the true input-output relationship is strictly linear (Appendix~\ref{app:linear}).

\subsection{Implementation Details}

The computation of the weighted \f mean $\mu(w)$ depends on the geometry of the output space and is carried out using space-specific numerical routines. We summarize below the implementations used in our numerical experiments.

\paragraph{One-dimensional probability distributions.}
For one-dimensional probability distributions equipped with the $2$-Wasserstein metric, the weighted \f mean admits a closed-form expression in terms of quantile functions. Specifically, $\mu(w)$ is computed as the distribution whose quantile function is given by the weighted average of the empirical quantile functions of the training distributions. This computation is exact and does not require iterative optimization.

\paragraph{Networks.}
For network-valued outputs represented by graph Laplacian matrices and equipped with the Frobenius metric, the weighted \f mean reduces to a weighted Euclidean average of the Laplacian matrices. In this case, $\mu(w)$ is computed directly by matrix summation and normalization; uniqueness of the \f mean is guaranteed.

\paragraph{SPD matrices (power metric).}
For SPD matrices equipped with the power metric, the weighted \f mean is computed via the power transform. Specifically, each matrix is first mapped to the power-transformed space, where the \f mean reduces to a weighted Euclidean average. The resulting matrix is then mapped back using the inverse power transform. This computation is direct, avoids
iterative optimization, and was numerically stable in all experiments.

\paragraph{SPD matrices (BW metric).}
For SPD matrices equipped with the BW metric, the underlying space is positively curved and the weighted \f mean need not be unique in general. In our implementation, we compute $\mu(w)$ using a standard fixed-point iteration for the BW barycenter \citep{alva:16}. The algorithm is initialized at the identity matrix and iteratively updates the current estimate via matrix square-root and inverse square-root operations obtained from eigendecomposition. The procedure returns one minimizer of the weighted \f functional. Although multiple minimizers may exist in principle, we did not observe numerical instability or non-convergence in our experiments. The empirical loss is therefore well-defined with respect to the selected barycenter.

Across all settings, \f mean computations are embedded within the training loop and are differentiable with respect to the weights $w$, enabling end-to-end optimization via backpropagation. The implementation details and code are publicly available at \url{https://github.com/SUIIAO/E2M}.

\subsection{Experimental Setup}
In both the numerical experiments and real-world data applications, E2M was trained for $2,000$ epochs using mini-batches of size $32$, a learning rate of $5\times 10^{-4}$, and a dropout rate of $30\%$. Other hyperparameters, including the regularization strength, the number of hidden layers and the number of neurons per layer, were selected via grid search based on cross-validated empirical risk (see Appendix \ref{app:hyper} for details). For each training run, $10\%$ of the training data was held out for early stopping. Performance was evaluated via mean squared prediction error (MSPE) over $200$ independent test points. For the $q$-th Monte Carlo run, with $\hat{m}_q$ denoting the estimator and $m$ the true regression function, the MSPE is
\[\mathrm{MSPE}_q = \frac{1}{200}\sum_{i=1}^{200} d^2\{\hat{m}_{q}(X_i^{\text{test}}), m(X_i^{\text{test}})\},\]
where $d$ is the metric for the corresponding metric space. The average performance over $200$ Monte Carlo runs is quantified by
\[\mathrm{AMSPE}=\frac{1}{200}\sum_{q=1}^{200}\mathrm{MSPE}_q.\]

\begin{figure}[tb]
    \centering
    \begin{subfigure}{0.45\textwidth}
        \centering
        \includegraphics[width=1\linewidth]{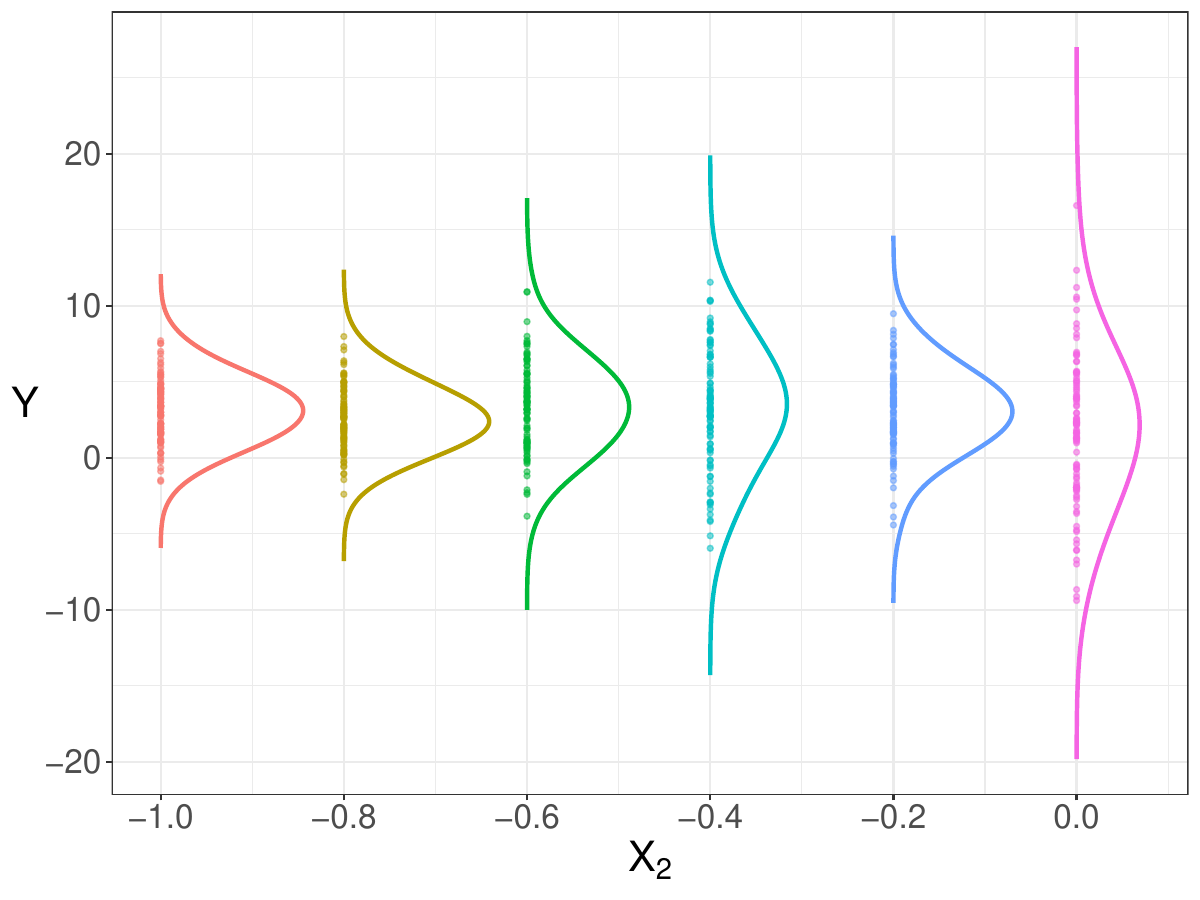}
        \caption{}
    \end{subfigure}\hfill
    \begin{subfigure}{0.45\textwidth}
        \centering
        \includegraphics[width=1\linewidth]{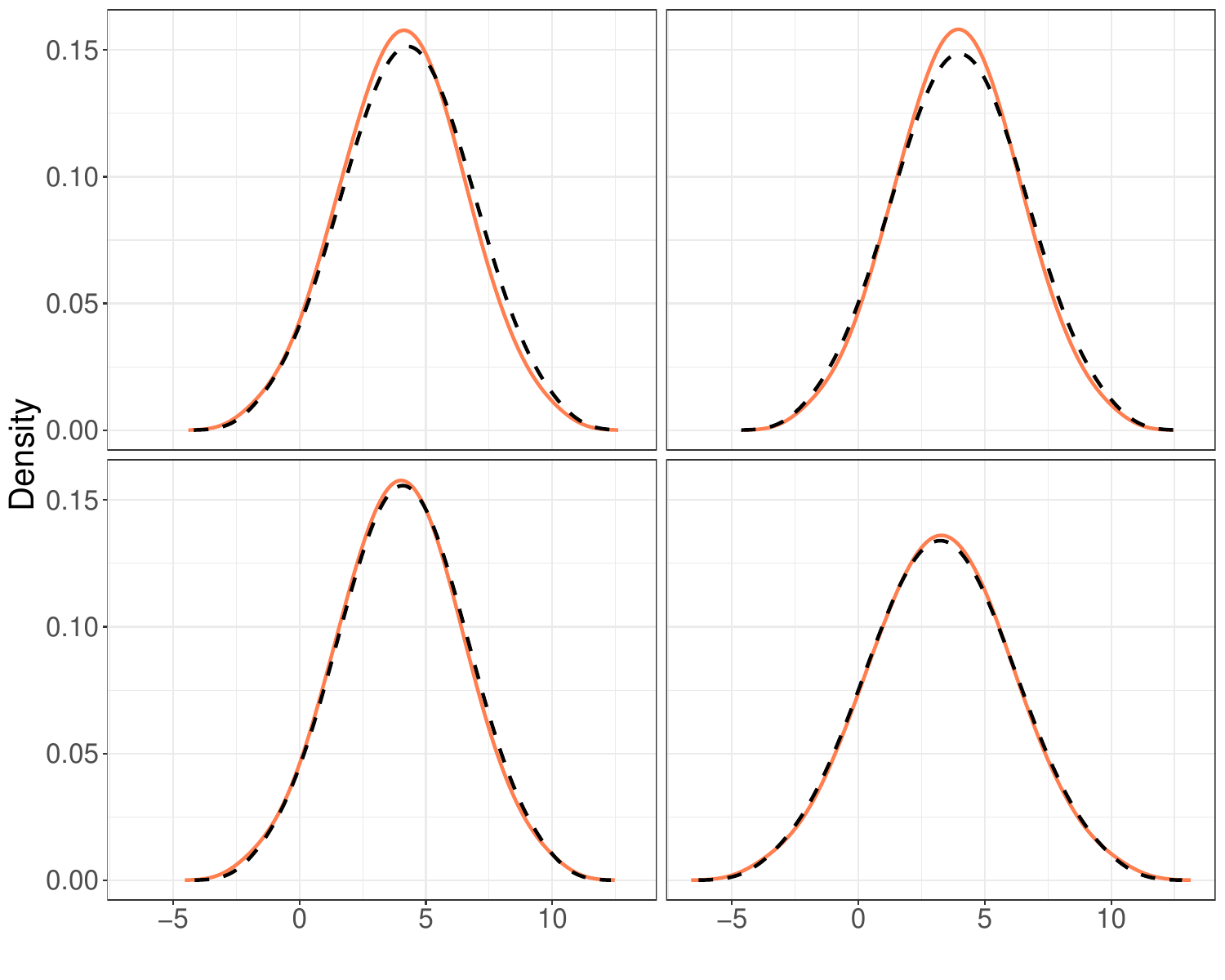}
        \caption{}
    \end{subfigure}
    \caption{(a) Simulated distributional outputs as a function of $X_2$, with all other covariates fixed at their mean (continuous variables) or median (discrete variables) values. (b) Fitted versus true densities in the distributional setting with $n=500$. Four randomly selected sample units are shown for illustration. Coral solid curves denote the E2M fitted densities; the black dashed curves denote the corresponding true densities.}
    \label{fig:dist_vis}
\end{figure}

\paragraph{Distributions.}
We consider a regression setting where the output $Y$ is a Gaussian distribution. The input is a vector $X \in \mathbb{R}^{12}$ with components generated as follows:
\begin{align*}
    &X_1 \sim U(-1,0),\quad\ X_2 \sim U(-1,0),\quad X_3 \sim U(0,1),\quad X_4 \sim U(0,1),\\
    &X_5 \sim \mathrm{Gamma}(2,2),\quad X_6 \sim \mathrm{Gamma}(3,2),\quad X_7 \sim \mathrm{Gamma}(4,2),\quad X_8 \sim \mathrm{Gamma}(5,2),\\
    &X_9 \sim \mathrm{Ber}(0.6),\quad X_{10} \sim \mathrm{Ber}(0.5),\quad X_{11} \sim \mathrm{Ber}(0.4),\quad X_{12}\sim \mathrm{Ber}(0.3).
\end{align*}
The mean $\eta$ and standard deviation $\sigma$ of the distributional output $Y$ are generated conditional on the input, where $\eta\sim N(\mu(X), 0.5^2), \sigma \sim\mathrm{Gamma}(\theta(X)^2, \theta(X)^{-1})$ with
\begin{align*}
    &\mu(X)=2 + 2\cos (\pi X_1)^2 + \sin(\pi X_2)^2 X_9 + \sqrt{X_5X_6}(1-X_9),\\
    &\theta(X)=1+\cos(\pi X_2/2) + \sin(\pi X_3)X_{10} + \sqrt{X_6X_7}(1- X_{10})/3.
\end{align*}
To better reflect real-world settings where the underlying probability distributions are not directly observable, we simulate independent samples from each distribution. Specifically, for each distribution $Y_i$, we generate 100 observations $\{y_{ij}\}_{j=1}^{100}$. E2M must then operate on a noisy version of $Y_i$, constructed from the empirical distribution of these samples; see also \citep{zhou:23}. To provide additional insight into the data-generating mechanism, Figure~\ref{fig:dist_vis} (a) visualizes simulated distributional outputs as a function of a representative covariate, $X_2$, while holding all remaining covariates fixed at their mean values (for continuous variables) or median levels (for binary variables). This plot illustrates the dependence of mean and variance of the distributional output $Y$ on $X_2$ implied by the model specification.

\paragraph{Networks.}
\label{app:network}
The output is a graph Laplacian derived from a weighted stochastic block model with two communities. Each network contains 10 nodes, equally divided into two blocks. The block connectivity structure is governed by probabilities $p_{11}=p_{22}=0.5$ for within-community connections and $p_{12}=0.2$ for between-community connections. The input is a vector $X\in\mathbb{R}^9$ with components generated as follows:
\begin{align*}
    &X_1 \sim U(0,1), \quad X_2 \sim U(-1/2,1/2), \quad X_3 \sim U(1,2),\quad
    X_4 \sim N(0,1),\\ &X_5 \sim N(0,1), \quad X_6 \sim N(5,5), \quad
    X_7 \sim \mathrm{Ber}(0.4), \quad X_8 \sim \mathrm{Ber}(0.3), \quad X_9 \sim \mathrm{Ber}(0.6).
\end{align*}
For each edge, we assign a Beta-distributed weight with shape parameters depending on $X$ and the block membership of the connected nodes. In Block 1, the shape parameters are $\alpha_1=2 \sin (\pi X_1) X_8+\cos (\pi X_2)(1-X_8)$ and $\beta_1=2 X_4^2 X_7+X_5^2(1-X_7)$. In Block 2, the parameters are $\alpha_3=\sin (\pi X_1) X_8+2 \cos (\pi X_2)(1-X_8)$ and $\beta_3=X_4^2 X_7+2 X_5^2(1-X_7)$. Between blocks, we set $\alpha_2=2 \sin (\pi X_1) X_8+\cos (\pi X_2)(1-X_8)$ and $\beta_2=X_4^2 X_7+2 X_5^2(1-X_7)$. These weights are assembled into an adjacency matrix and the corresponding graph Laplacian serves as the output.

\paragraph{SPD matrices with the power metric.}
The third simulation generates SPD matrix outputs from a Wishart distribution, $Y \sim \mathcal{W}_l(\Sigma, df)$, where $l=5$ is the dimension of the matrices, $df = l+1$ is the degrees of freedom, and $\Sigma$ is the scale matrix with input-dependent diagonal entries. The input is a vector $X\in\mathbb{R}^{12}$ with components generated as follows:
\begin{align*}
    &X_1 \sim U(0,1), \quad X_2 \sim U\left(-\frac{1}{2}, \frac{1}{2}\right), \quad X_3 \sim U(1,2),\quad X_4 \sim \mathrm{Gamma}(3,2), \\& X_5 \sim \mathrm{Gamma}(4,2),\quad X_6 \sim \mathrm{Gamma}(5,2), \quad X_7 \sim N(0,1), \quad X_8 \sim N(0,1), \\& X_9 \sim N(0,1), \quad X_{10} \sim \mathrm{Ber}(0.4), \quad X_{11} \sim \mathrm{Ber}(0.5), \quad X_{12} \sim \mathrm{Ber}(0.6).
\end{align*}
The diagonal entries of the scale matrix $\Sigma$ are generated conditional on the input, where
\begin{align*}
    &\Sigma_{11} = \{\sin(\pi X_1) X_{10} + \cos(\pi X_2) (1 - X_{10})\}^2,\quad \Sigma_{22} = \sin^2(\pi X_1) \cos^2(\pi X_2), \\
    &\Sigma_{33} = \{\frac{X_4}{X_5} \cdot \frac{1}{10} X_{11} + \sqrt{\frac{X_5}{X_4}} \cdot \frac{1}{10} (1 - X_{11})\}^2,\quad \Sigma_{44} = \frac{|X_7 X_8|}{25},\quad \Sigma_{55} = \frac{|X_9 / X_6|}{9}.
\end{align*}
Distances between SPD matrices are computed using the power metric with exponent 1/2 \citep{dryd:09}.

\paragraph{SPD matrices with the BW metric.}
To further investigate the applicability of E2M beyond the Hadamard setting, we conducted an additional experiment with SPD matrices equipped with the BW metric, a classical example of a positively curved space that is not Hadamard \citep{than:23}. The BW distance between two SPD matrices $A, B \in \mathrm{Sym}_l^+$ is given by
\[d^2_{\mathrm{BW}}(A,B) = \mathrm{Tr}(A) + \mathrm{Tr}(B) - 2\,\mathrm{Tr}\!\left( (A^{1/2} B A^{1/2})^{1/2} \right).\]

We demonstrate this setting with $2 \times 2$ SPD outputs, while noting that the same implementation readily extends to higher dimensions. For each sample, predictors $X=(X_1,\ldots,X_5)\in\mathbb{R}^5$ were generated as
\[X_1 \sim U(0,1), \quad X_2 \sim U(-0.5,0.5), \quad X_3 \sim U(1,2), \quad X_4 \sim \mathrm{Ber}(0.6), \quad X_5 \sim \mathrm{Ber}(0.5).\]
Each SPD output $Y$ was drawn from a Wishart distribution with degrees of freedom $df=3$ and scale matrix $\Sigma(X)=\mathrm{diag}(\sigma_{11}^2,\sigma_{22}^2)$, where
\[\sigma_{11} = \sin(\pi X_1)X_4 + \cos(\pi X_2)(1-X_4), \quad \sigma_{22} = \sin(\pi X_2)\cos(\pi X_3).\]

\subsection{Discussion of the Simulation Results}
Table~\ref{tab:sim} summarizes the predictive performance of E2M and baseline methods across all simulation settings. For distributional, network  and SPD (power metric) outputs, E2M consistently achieves the lowest average prediction error among all competing methods, with its advantage becoming more pronounced at larger sample sizes. For SPD outputs with the BW metric, which correspond to a positively curved space outside the Hadamard class, E2M remains fully implementable and delivers low prediction error, whereas DFR, GFR, RFWLFR, SDR and IFR cannot currently handle this setting. These findings demonstrate that E2M effectively models complex nonlinear relationships between inputs and a wide range of non-Euclidean outputs, encompassing both Hadamard and non-Hadamard spaces.

\begin{table}[tb]
\centering
\caption{Average mean squared prediction errors (mean on first line, standard deviation in parentheses on second line) of E2M, deep \f regression (DFR) \citep{iao:25}, global \f regression (GFR) \citep{mull:19:6}, sufficient dimension reduction (SDR) \citep{zhan:21:1}, single index \f regression (IFR) \citep{mull:23:3}  and random forest weighted local \f regression (RFWLFR) \citep{qiu:24} for distribution, network  and SPD matrix outputs. SDR and IFR were not included for SPD outputs with the power metric,  due to the lack of available implementations, and none of the other competing methods currently support SPD outputs under the BW metric.}
\label{tab:sim}
\begin{tabular}{cc|cccccc}
\toprule
Output & $n$ & E2M & DFR & GFR & RFWLFR & SDR & IFR \\
\midrule
\multirow{8}{*}{Distribution}
& 500  & \textbf{0.562} & 0.869 & 0.766 & 0.927 & 0.753 & 0.929 \\
&      & (0.120) & (0.125) & (0.058) & (0.110) & (0.105) & (0.078) \\
& 1000 & \textbf{0.415} & 0.541 & 0.742 & 0.811 & 0.660 & 0.933  \\
&      & (0.058) & (0.189) & (0.055) & (0.087) & (0.080) & (0.074) \\
& 2000 & \textbf{0.218} & 0.295 & 0.729 & 0.691 & 0.623 & 0.930 \\
&      & (0.048) & (0.093) & (0.049) & (0.065) & (0.064) & (0.071) \\
\midrule
\multirow{8}{*}{Network}
& 500  & \textbf{4.672} & 7.114 & 9.901 & 7.278 & 7.049 & 9.792 \\
&      & (0.983) & (1.108) & (0.623) & (0.766) & (0.770) & (0.657) \\
& 1000 & \textbf{2.849} & 4.565 & 9.683 & 5.607 & 6.580 & 9.622 \\
&      & (0.623) & (0.750) & (0.570) & (0.544) & (0.698) & (0.642) \\
% & 1500 & \textbf{2.076} & 3.604 & 9.560 & 6.355 & 9.535 \\
% &      & (0.403)        & (0.557) & (0.611) & (0.675) & (0.685) \\
& 2000 & \textbf{1.729} & 3.018 & 9.582 & 4.430 & 6.403 & 9.561 \\
&      & (0.381) & (0.515) & (0.583) & (0.393) & (0.679) & (0.696) \\
\midrule
\multirow{8}{*}{\makecell{SPD matrix\\(power metric)}}
& 500  & \textbf{0.443} & 1.084 & 1.118 & 0.865 & --- & --- \\
&      &  (0.090) & (0.283) & (0.076) &   (0.093) &  &     \\
& 1000 & \textbf{0.279} & 0.582 & 1.099 & 0.726 & --- & --- \\
&    &  (0.045) & (0.103) & (0.068) &  (0.080) &  &     \\
% & 1500 & & & & --- & --- \\
% &      & & & & --- & --- \\
& 2000 & \textbf{0.187} & 0.346 & 1.083 & 0.637 & --- & --- \\
&      & (0.034) & (0.039) & (0.057) &  (0.067) &   &     \\
\midrule
\multirow{8}{*}{\makecell{SPD matrix\\(BW metric)}}
& 500  & \textbf{0.342} & --- & --- & --- & --- & ---\\
&      & (0.029)        &     &     &     &     & \\
& 1000 & \textbf{0.288} & --- & --- & --- & --- & --- \\
&      & (0.026)        &     &     &     &     & \\
% & 1500 & \textbf{}      & --- & --- & --- & --- \\
% &      & ()             &     &     &     &     \\
& 2000 & \textbf{0.250}      & --- & --- & --- & --- & --- \\
&      & (0.025)             &     &     &     &   &   \\
\bottomrule
\end{tabular}
\end{table}

The distributional outputs in our simulations correspond to Gaussian distributions characterized by their mean and standard deviation and therefore lie on a two-dimensional manifold embedded in the Wasserstein space. This structure is highly aligned with the assumptions of DFR, which is specifically designed to exploit low-dimensional geometry in the output space. Despite this favorable setting, E2M consistently outperforms DFR across all sample sizes, achieving at least a $23\%$ reduction in prediction error. This result further underscores the flexibility and effectiveness of E2M, even in scenarios where competing methods are specifically adapted to handle the underlying data structure.

To complement the MSPE comparisons, Figure~\ref{fig:dist_vis} (b) provides representative visualizations of fitted and true distributional outputs for four randomly selected sample units in the case of $n=500$. The fitted distributions produced by E2M closely track both the location and dispersion of the corresponding true distributions.

\subsection{Scalability via Anchor-Based Strategy}\label{subsec:anchor}
To evaluate scalability, we conducted an additional experiment using the anchor-based strategy introduced in Remark~\ref{rem:anchor}. Competing methods SDR and IFR become impractical beyond $n=2000$, requiring more than 20 minutes and one hour per run, respectively, and DFR is similarly limited due to the need to compute large pairwise distance matrices and run Dijkstra's algorithm. For this reason, we focused on comparing E2M against GFR at larger scales.

Instead of using all training outputs as anchors for computing the weighted \f mean, we fixed a random subset of 1000 outputs as anchors, which makes the optimization complexity independent of the total sample size. Following the same simulation setup as before for both distributional and network outputs, we considered a large-scale setting with $n=10{,}000$ and compared E2M against GFR over 200 Monte Carlo replications.

Table~\ref{tab:anchor} reports the results for $n=10{,}000$, while results for $n=500,1000,2000$ can be found in Table~\ref{tab:sim} for comparison. E2M scales efficiently: predictive accuracy continues to improve with $n$  and training with $n=10{,}000$ completed in about 5 minutes on a standard laptop, only slightly longer than the 4 minutes required at $n=2000$ when using all outputs as anchors. These findings confirm that the anchor-based strategy provides strong scalability while preserving predictive performance.

\begin{table}[tb]
\centering
\caption{Average mean squared prediction errors and standard deviations (in parentheses) for distributional and network outputs at $n=10{,}000$ using the anchor-based strategy.}
\label{tab:anchor}
\vskip 0.1in
\begin{tabular}{c|cc}
\toprule
Output & E2M & GFR \\
\midrule
Distribution & \textbf{0.088} (0.017) & 0.717 (0.048) \\
Network      & \textbf{1.072} (0.356) & 9.416 (0.565) \\
\bottomrule
\end{tabular}
\end{table}

\section{Data Applications}
We evaluate the effectiveness of E2M on two real-world regression tasks: modeling age-at-death distributions from international human mortality data and predicting daily transportation networks from New York City yellow taxi data. Both applications involve complex non-Euclidean outputs, probability distributions and networks, that cannot be adequately modeled using classical regression techniques.

\subsection{Human Mortality Data}
\label{sec:mortality}
We analyze age-at-death distributions across 162 countries in the year 2015 using life table data published by the United Nations World Population Prospects 2024 (\url{https://population.un.org/wpp/downloads}). For each country and age group, the life table reports the number of deaths aggregated in five-year age intervals, forming histograms with uniform bin width. Using the \texttt{frechet} package \citep{chen:20}, we apply local linear smoothing to these histograms and standardize them using trapezoidal integration, yielding continuous probability density functions that serve as regression outputs. Each country is associated with a nine-dimensional predictor vector comprising demographic, economic and environmental indicators, listed in Table~\ref{tab:pre_mor}. These predictors capture key aspects of socioeconomic conditions, such as population density, GDP per capita and healthcare expenditures, which are known to influence life expectancy and mortality patterns.

\begin{table}[tb]
\centering
\caption{Predictors of human mortality data.}
\label{tab:pre_mor}
\begin{tabular}{p{0.15\linewidth} | p{0.3\linewidth} | p{0.45\linewidth}}
\toprule
Category & Predictor & Explanation \\
\midrule
\multirow{5}{*}{Demography} & 1. Population Density & population per square kilometer \\
\cmidrule{2-3}
& \multirow{2}{*}{2. Sex Ratio} & number of males per 100 females in the population \\
\cmidrule{2-3}
& \multirow{2}{*}{3. Mean Childbearing Age} & average age of mothers at the birth of their children \\
\midrule
\multirow{10}{*}{Economics} & 4. GDP & gross domestic product per capita \\
\cmidrule{2-3}
& \multirow{2}{*}{5. GVA by Agriculture} & percentage of agriculture, hunting, forestry and fishing activities of gross value added \\
\cmidrule{2-3}
 & \multirow{2}{*}{6. CPI} & consumer price index treating 2010 as the base year\\
\cmidrule{2-3}
& \multirow{2}{*}{7. Unemployment Rate} & percentage of unemployed people in the labor force\\
\cmidrule{2-3}
& \multirow{1}{*}{8. Health Expenditure} & percentage of expenditure on health of GDP\\
\midrule
Environment & 9. Arable Land & percentage of total land area \\
\bottomrule
\end{tabular}
\end{table}

\begin{figure}[tb]
    \centering
    \begin{subfigure}{0.45\textwidth}
        \centering
        \includegraphics[width=1\linewidth]{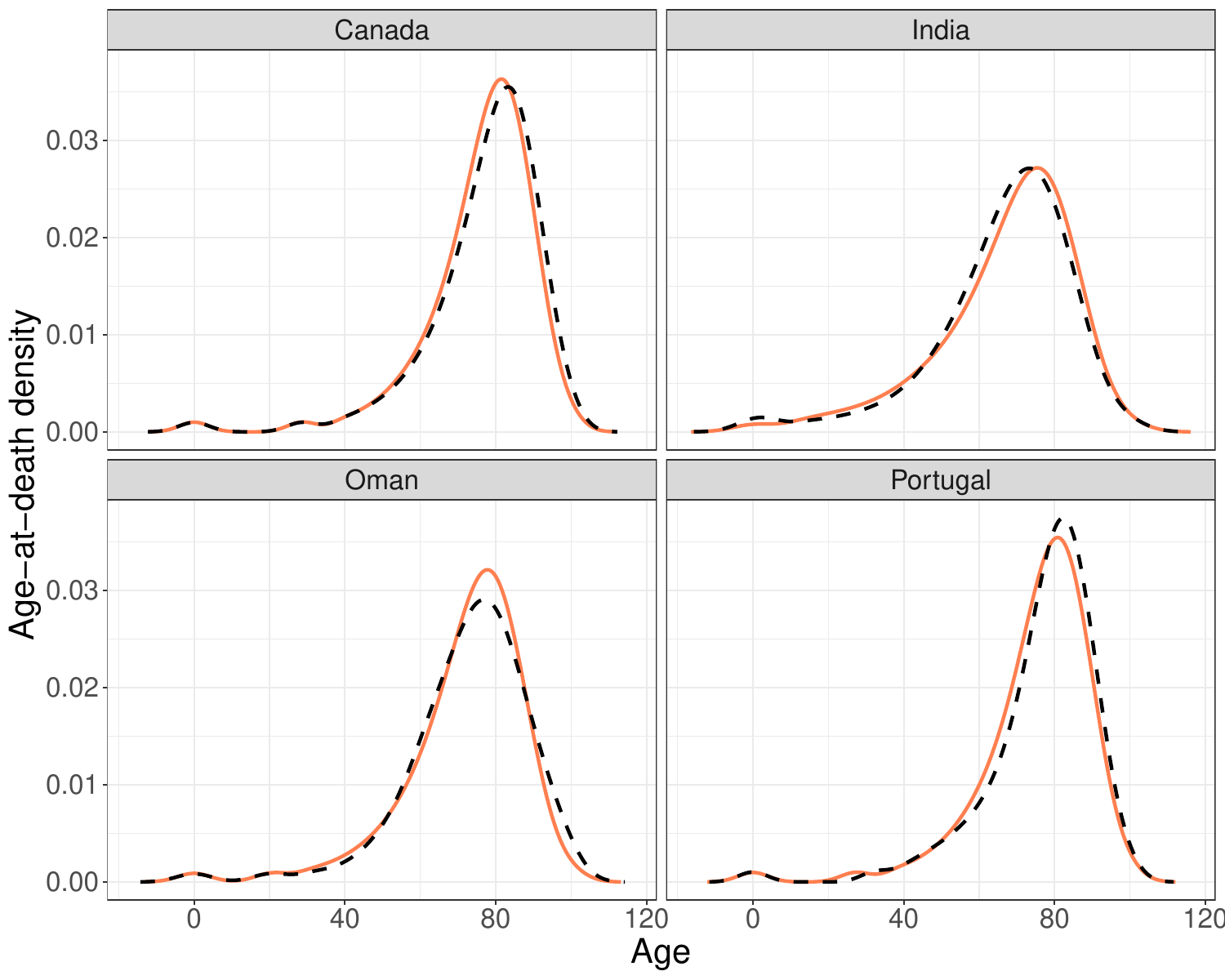}
        \caption{}
    \end{subfigure}\hfill
    \begin{subfigure}{0.45\textwidth}
        \centering
        \includegraphics[width=1\linewidth]{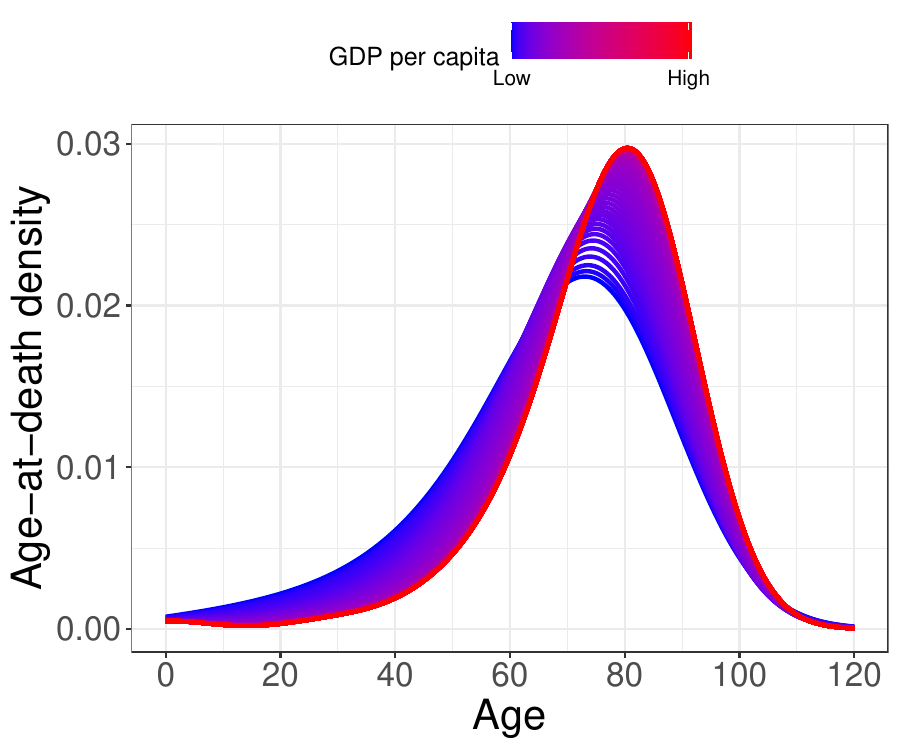}
        \caption{}
    \end{subfigure}
    \caption{(a) Comparison between observed and predicted age-at-death densities for four countries: Canada, Portugal, Oman and India. The coral solid curves denote the densities estimated by E2M; the black dashed curves denote the corresponding observed empirical densities. (b) Age-at-death densities at different levels of GDP.}
    \label{fig:mor}
\end{figure}

Predictive performance is assessed via leave-one-out cross-validation, with MSPE as the evaluation criterion. Table~\ref{tab:app} reports the results, showing that E2M achieves the lowest MSPE among all methods considered. Despite the modest sample size, E2M demonstrates clear gains over competing approaches, underscoring its ability to capture nonlinear relationships between country-level covariates and complex distributional outcomes. To further illustrate prediction quality, Figure~\ref{fig:mor} (a) compares the observed and fitted age-at-death densities for four countries (Canada, Portugal, Oman and India). The fitted densities closely track the main structural features of the observed distributions, providing qualitative evidence of the model's adequacy in the $n=162$ setting. To explore covariate effects, Figure~\ref{fig:mor} (b) illustrates the predicted age-at-death densities obtained by varying GDP per capita from low to high values while holding all other predictors fixed at their median levels; the resulting rightward shift of the distributions indicates increasing life expectancy as GDP rises.

\begin{table}[tb]
\centering
\caption{Average mean squared prediction errors and standard deviations (in parentheses) of E2M, deep \f regression (DFR) \citep{iao:25}, global \f regression (GFR) \citep{mull:19:6}, sufficient dimension reduction (SDR) \citep{zhan:21:1} and single index \f regression (IFR) \citep{mull:23:3} for human mortality and taxi network data.}
\label{tab:app}
\begin{tabular}{c|ccccc}
\toprule
Data & E2M & DFR & GFR & SDR & IFR \\
\midrule
\multirow{2}{*}{Human mortality}
& \textbf{22.64} & 26.75 & 31.32 & 27.60 & 42.57 \\
& (41.32)        & (51.19) & (58.83) & (44.40) & (76.22) \\
\midrule
\multirow{2}{*}{Taxi network}
& \textbf{6.83} & 7.93 & 12.40 & 13.38 & 42.66 \\
& (0.41)        & (0.50) & (0.18) & (0.30) & (5.56) \\
\bottomrule
\end{tabular}
\end{table}

\subsection{New York City Yellow Taxi Data}
\label{sec:taxi}
We study daily transportation patterns in Manhattan using yellow taxi trip records released by the New York City Taxi and Limousine Commission (\url{https://www.nyc.gov/site/tlc/about/tlc-trip-record-data.page}). These records include detailed information such as pick-up and drop-off locations, trip distances, passenger counts, fares and payment methods. To capture external influences, we also collect daily weather history from Weather Underground (\url{https://www.wunderground.com/history/daily/us/ny/new-york-city/KLGA/date}), including temperature, humidity, wind speed, pressure and precipitation.

Following the preprocessing steps described in  \citet{mull:22:11}, the 66 original taxi zones were  grouped into 13 regions. For each day between January 1, 2018 and December 31, 2019, we constructed  a weighted directed network where nodes represent regions and edge weights denote the number of passengers traveling between region pairs. Each network is represented by a $13 \times 13$ graph Laplacian matrix and paired with a 13-dimensional predictor vector comprising daily weather information, calendar effects (e.g., day-of-week indicators) and aggregated trip statistics. A full list of predictors is shown in Table~\ref{tab:predictor}.

To evaluate predictive performance, ten-fold cross-validation was repeated across 100 Monte Carlo runs, computing MSPE for each method. Results in Table~\ref{tab:app} show that E2M achieves the lowest prediction error, outperforming all competing methods by a substantial margin. By explicitly respecting the geometry of network outputs, E2M is able to capture complex dependencies in daily passenger flows that methods relying on embeddings or restrictive assumptions fail to model adequately.

\begin{table}[tb]
\centering
\caption{Predictors of New York City taxi network data.}
\label{tab:predictor}
\begin{tabular}{p{0.15\linewidth} | p{0.3\linewidth} | p{0.45\linewidth}}
\toprule
Category & Predictor & Explanation \\
\midrule
\multirow{5}{*}{Weather} & 1. Temp & daily average temperature \\
& 2. Humidity & daily average humidity \\
& 3. Wind & daily average windspeed \\
& 4. Pressure & daily average barometric pressure \\
& 5. Precipitation & daily total precipitation \\
\midrule
Year & 6. Year & indicator for the year of 2018 \\
\midrule
\multirow{2}{*}{Day} & 7. Mon to Thur & indicator for Monday to Thursday\\
& 8. Friday or Saturday & indicator for Friday or Saturday\\
\midrule
\multirow{5}{*}{Trip}& 9. Passenger Count & daily average number of passengers \\
& 10. Trip Distance & daily average trip distance \\
& 11. Fare Amount & daily average fare amount \\
& 12. Tip Amount & daily average tip amount \\
& 13. Tolls Amount & daily average tolls amount\\
\bottomrule
\end{tabular}
\end{table}

\section{Discussion}
This paper introduces E2M, a novel end-to-end regression framework for metric space-valued outputs that fully exploits the representational capacity of deep neural networks while respecting the geometry of the output space. By incorporating entropy regularization into the learned weight distribution, E2M enables a data-driven trade-off between localized regression and global smoothing. We establish a universal approximation theorem that demonstrates the expressive power of E2M for approximating conditional \f means and analyze the algorithmic convergence for the proposed training algorithm. Empirically, E2M achieves superior performance across diverse simulated and real-world datasets with complex metric space-valued outputs.

From a learning theoretic perspective, prior work on prediction in metric spaces has primarily focused on characterizing when Bayes consistent prediction is possible and on constructing nonparametric, data-supported rules that achieve such guarantees. Existing results address classification with metric inputs and discrete labels \citep{hann:21} or regression under general metric losses using fixed aggregation schemes such as nearest-neighbor or medoid-based rules \citep{cohe:22}. In contrast, E2M adopts a complementary viewpoint centered on expressive modeling and optimization: predictions are defined implicitly through learned, input-dependent weighting mechanisms parameterized by neural networks. Extending learning-theoretic guarantees, such as Bayes consistency or finite-sample risk bounds, to such end-to-end geometry-aware models remains an important direction for future work.

Beyond prediction, an important practical question concerns the assessment of covariate effects and their uncertainty. In Appendix~\ref{app:vi}, we provide a permutation-based variable importance procedure adapted to the \f regression setting. This model-agnostic approach quantifies the predictive relevance of each covariate by measuring performance degradation under permutation, offering a practical tool for exploratory analysis in complex metric space-valued models. However, this procedure does not constitute formal hypothesis testing in the classical regression sense. Developing principled methods for statistical inference and uncertainty quantification in \f regression remains technically challenging due to the nonlinear geometry of the output space and the implicit definition of the \f mean. Constructing confidence regions, testing covariate effects and deriving asymptotic distributions for learned predictors are important open problems that warrant further investigation.

Several additional limitations suggest future research directions. First, E2M assumes access to a well defined metric on the output space. In practice, multiple plausible metrics may exist, each inducing different geometric and statistical properties; the choice of metric can substantially influence model behavior. Integrating metric learning techniques \citep{xing:02, kaya:19} with geometry-aware regression may provide a principled way to tailor the metric to the data and task at hand. Second, although many metric spaces of practical interest are Hadamard spaces, part of our theoretical analysis, in particular the convergence guarantees for the training algorithm, relies on this assumption. Extending the theoretical framework to broader classes of metric spaces would further enhance the generality of the method. Finally, a promising extension is to settings where both inputs and outputs lie in general metric spaces. Combining geometry-aware output modeling as in E2M with recent advances in geometric deep learning for non-Euclidean inputs \citep{bron:17} could enable flexible regression frameworks for tasks such as network to network prediction or mapping between structured geometric objects.

% Acknowledgements and Disclosure of Funding should go at the end, before appendices and references

\acks{We wish to thank the editor and the reviewers for their valuable comments and suggestions, which have significantly improved the quality of the paper. This research was partially supported by NSF grant DMS-2310450. }


% \newpage

\appendix
% Note: in this sample, the section number is hard-coded in. Following
% proper LaTeX conventions, it should properly be coded as a reference:

\section{Characterizations of Mean and Conditional Mean}
\label{app:mean}
Let $Y\in\mathbb{R}$ be a real-valued random variable with finite second moment. Then, the expectation of $Y$ can be equivalently characterized as the minimizer of the expected squared deviation:
\[E[Y] = \argmin_{y\in\mathbb{R}} E[(Y-y)^2].\]
This identity follows by expanding the square and minimizing
\[E[(Y - y)^2] = E[Y^2] - 2y E[Y] + y^2.\]
Solving
\[\frac{d}{dy} E[(Y - y)^2] = 0\]
yields $y = E[Y]$ as the unique minimizer.

For a random pair $(X, Y) \in \mathbb{R}^p \times \mathbb{R}$, the conditional expectation of $Y$ given $X = x$ can be similarly expressed as
\[E[Y | X = x] = \argmin_{y \in \mathbb{R}} E[(Y - y)^2 | X = x].\]

These characterizations naturally extend to random objects taking values in a general metric space $(\Omega, d)$. When $Y \in \Omega$, the \f mean \citep{frec:48} is defined as the minimizer of the expected squared distance,
\[E_\oplus[Y] = \argmin_{y \in \Omega} E[d^2(y, Y)].\]
For a random pair $(X, Y) \in \mathbb{R}^p \times \Omega$, the conditional \f mean \citep{mull:19:6} is given by
\[E_\oplus[Y | X = x] = \argmin_{y \in \Omega} E[d^2(y, Y) | X = x].\]
When $\Omega = \mathbb{R}$ and $d(y,Y)=|y-Y|$, these definitions recover the classical mean and conditional mean.

\section{Characterization of Regression via Weighted Average}
\label{app:reg}
The variational view of conditional expectation leads to a natural formulation of regression as the minimization of weighted squared loss. In linear regression, the regression function is assumed to be
\begin{equation}\label{eq:linear}
    m(x) = E[Y | X = x] = \beta_0 + \beta_1^\top(x - \mu),
\end{equation}
where $\mu = E[X]$. The coefficients $\beta_0$ and $\beta_1$ are chosen to minimize the expected squared residual:
\[(\beta_0, \beta_1) = \argmin_{\beta_0, \beta_1} E_{X} \left[ E_{Y \mid X} \left\{ \left(Y - \beta_0 - \beta_1^\top(X - \mu)\right)^2 \right\} \right].\]
Letting $\Sigma_{XX} = \mathrm{Cov}(X)$ and $\Sigma_{XY} = E[(X - \mu)Y]$, the optimal coefficients are
\[\beta_0 = E[Y], \quad \beta_1 = \Sigma_{XX}^{-1} \Sigma_{XY}.\]
Substituting these into \eqref{eq:linear} gives
\begin{align*}
    m(x) &= E[Y] + \Sigma_{XY}^\top \Sigma_{XX}^{-1}(x - \mu) \\
         &= E\left[ Y + Y (X - \mu)^\top \Sigma_{XX}^{-1} (x - \mu) \right] \\
         &= E[w(x; X) Y],
\end{align*}
with weight function
\[w(x; X) = 1 + (X - \mu)^\top \Sigma^{-1}(x - \mu).\]
Since $E[w(x; X)] = 1$, we may express the regression function as the minimizer of a weighted squared deviation:
\begin{align*}
    m(x) & = \argmin_{y\in\mathbb{R}}\{y - E[w(x;X)Y]\}^2\\
    & = \argmin_{y\in\mathbb{R}} \{y^2 - 2 y E[w(x;X)Y]\}\\
    & = \argmin_{y\in\mathbb{R}} E[y^2 w(x;X) - 2 y w(x;X)Y + w(x;X) Y^2]\\
    & = \argmin_{y\in\mathbb{R}} E[w(x;X) (Y - y)^2].
\end{align*}
This formulation reveals that linear regression solves a weighted least squares problem, where weights reflect the alignment between input and the target point $x$.

This perspective also applies to local linear regression. For simplicity, we consider scalar predictors $X \in \mathbb{R}$. The local linear estimator \citep{fan:96} $m(x) = \beta_0(x)$ is defined via a locally weighted least squares problem:
\begin{equation}\label{eq:llr}
    (\beta_0, \beta_1) = \argmin_{\beta_0, \beta_1} E\left[ K_h\left(X - x\right)\left\{Y - \beta_0 - \beta_1\left(X - x\right)\right\}^2\right],
\end{equation}
where $K_h$ is a kernel function with bandwidth $h$.

Writing  $\mu_j = E[K_h(X - x)(X - x)^j]$, $r_j = E[K_h(X - x)(X - x)^j Y]$ and $\sigma_0^2 = \mu_0 \mu_2 - \mu_1^2$, the solutions to \eqref{eq:llr} are
\[\beta_0(x) = \frac{\mu_2 r_0 - \mu_1 r_1}{\sigma_0^2}, \quad \beta_1(x) = \frac{\mu_0 r_1 - \mu_1 r_0}{\sigma_0^2}.\]
Therefore, the local linear regression function is
\begin{align*}
    m(x) & = \beta_0(x) = \frac{\mu_2 r_0 - \mu_1 r_1}{\sigma_0^2}\\
     & = \frac{1}{\sigma_0^2} E[\mu_2 K_h(X - x) Y - \mu_1 K_h(X - x)(X - x) Y]\\
     & = E[ w(x; X) Y ],
\end{align*}
where the weight function is given by
\[w(x; X) = \frac{K_h(X - x) (\mu_2 - \mu_1 (X - x))}{\sigma_0^2}.\]
Observe that
\begin{align*}
    E[w(x; X)] &= E[\frac{K_h(X - x) (\mu_2 - \mu_1 (X - x))}{\sigma_0^2}]\\
    &= \frac{\mu_2\mu_0 - \mu_1^2}{\sigma_0^2} \\
    & = 1.
\end{align*}
Similarly to linear regression, the local linear regression function can be alternatively represented as the minimizer of a weighted squared deviation:
\begin{align*}
    m(x) & = \argmin_{y\in\mathbb{R}} \{y - E[w(x;X)Y]\}^2\\
    & = \argmin_{y\in\mathbb{R}} E[w(x;X) (Y - y)^2].
\end{align*}
The local linear estimator can therefore be viewed as solving a localized version of the weighted least squares problem, where weights adapt to the target $x$ through a kernel mechanism.

These characterizations establish a unifying framework in which both linear and local linear regression estimate the conditional mean through weighted minimization of squared deviations. The nature of the weight function $w(x; X)$, whether globally defined or locally adaptive, determines the behavior and flexibility of the estimator. This weighting perspective motivated the development of \f regression \citep{mull:19:6}, which generalizes classical regression techniques to settings with metric space-valued outputs. Specifically, linear regression extends naturally to global \f regression, while local linear regression corresponds to local \f regression. These extensions are achieved by replacing Euclidean distances with general metric distances, thereby preserving the interpretation of regression as a weighted deviation minimization. The resulting estimators retain the structural intuition of their classical counterparts while enabling flexible modeling of complex, structured data.

\section{Proofs}\label{app:proof}
\subsection{Proof of Lemma~\ref{lem:frechet-continuity}}
\begin{proof}
We apply the Berge Maximum Theorem \cite[Theorem 17.31]{alip:06} to the weighted \f mean minimization problem. For completeness, we restate the theorem below.
\\

\noindent\textbf{Berge Maximum Theorem \citep[Theorem 17.31]{alip:06}} Let $\varphi: X \to Y$ be a continuous correspondence between topological spaces with nonempty compact values. Suppose the function $f: \mathrm{Gr}\, \varphi \mapsto \mathbb{R}$ is continuous, where $\mathrm{Gr}\, \varphi = \{ (x, y) \in X \times Y | y \in \varphi(x) \}$ denotes the graph of $\varphi$. Define the ``value function'' $m:X\mapsto\mathbb{R}$ by
\[
m(x) = \max_{y \in \varphi(x)} f(x, y),
\]
and the correspondence $\mu:X\to Y$ of maximizers by
\[
\mu(x) = \{ y \in \varphi(x) | f(x, y) = m(x) \}.
\]
Then:
\begin{enumerate}
    \item The value function $m$ is continuous.
    \item The ``argmax'' correspondence $\mu$ has nonempty compact values.
    \item If either $f$ has a continuous extension to all of $X \times Y$ or $Y$ is Hausdorff, then the ``argmax'' correspondence $\mu$ is upper hemicontinuous.
\end{enumerate}

We now match our setting to the theorem. Let $X = \Delta^{n-1}$, the $(n-1)$-simplex and $Y = \Omega$, the ambient metric space. Define the correspondence $\varphi: \Delta^{n-1} \to \Omega$ by $\varphi(w) = \Omega$ for all $w \in \Delta^{n-1}$, so that $\mathrm{Gr}\, \varphi = \Delta^{n-1} \times \Omega$. Since $\Omega$ is compact and Hausdorff, $\varphi(w)$ has nonempty compact values and the graph is well-defined.

We define the objective function $f: \Delta^{n-1} \times \Omega \mapsto \mathbb{R}$ by
\[
f(w, y) = -\sum_{i=1}^n w_i d^2(y, Y_i).
\]
Here we include a minus sign so that minimizing $\sum_{i=1}^n w_i d^2(y, Y_i)$ becomes equivalent to maximizing $f(w, y)$, as required for the application of the Berge Maximum Theorem.

We now verify the continuity of $f$. For fixed $y\in\Omega$, $w \mapsto f(w, y)$ is linear, hence continuous. Next, fix $w \in \Delta^{n-1}$ and consider continuity in $y$. For any $y_1, y_2 \in \Omega$, we have
\[
\begin{aligned}
|f(w, y_2) - f(w, y_1)| &= \left| \sum_{i=1}^n w_i \left( d^2(y_2, Y_i) - d^2(y_1, Y_i) \right) \right| \\
&\leq \sum_{i=1}^n w_i \left| d^2(y_2, Y_i) - d^2(y_1, Y_i) \right|.
\end{aligned}
\]
Using the identity
\[
d^2(y_2, Y_i) - d^2(y_1, Y_i) = (d(y_2, Y_i) + d(y_1, Y_i))(d(y_2, Y_i) - d(y_1, Y_i)),
\]
and applying the triangle inequality for the metric $d$, we obtain
\[
\left| d^2(y_2, Y_i) - d^2(y_1, Y_i) \right| \leq 2Dd(y_2, y_1),
\]
where $D = \sup_{u,v \in \Omega}d(u,v)$ is the diameter of $\Omega$ and is finite by compactness.

Thus,
\[
|f(w, y_2) - f(w, y_1)| \leq 2Dd(y_2, y_1) \sum_{i=1}^n w_i = 2Dd(y_2, y_1),
\]
since $\sum_{i=1}^n w_i = 1$. Therefore, $f(w, \cdot)$ is Lipschitz continuous with constant $2D$ in $y$ and hence continuous.

Since $f$ is continuous in both $w$ and $y$, and $\mathrm{Gr}\, \varphi = \Delta^{n-1} \times \Omega$ is compact, $f$ is jointly continuous on $\mathrm{Gr}\, \varphi$ and we conclude that all conditions of the Berge Maximum Theorem are satisfied. It follows that the value function
\[
m(w) = \max_{y \in \Omega} f(w, y)
\]
is continuous, and that the correspondence
\[
\mu(w) = \{ y \in \Omega | f(w, y) = m(w) \}
\]
has nonempty compact values and is upper hemicontinuous.

Under Assumption~\ref{ass:uniq}, the weighted \f mean $\mu(w)$ is unique for each $w \in \Delta^{n-1}$. Hence $\mu: \Delta^{n-1} \to \Omega$ is singleton-valued, and for singleton-valued correspondences, upper hemicontinuity implies continuity. Therefore, $\mu$ is continuous on $\Delta^{n-1}$.
\end{proof}

\subsection{Proof of Theorem~\ref{thm:universal-approximation}}
\begin{proof}
We aim to show that there exists a neural network $w_{\theta^*}$ such that the composed function $m_{\theta^*} = \mu \circ w_{\theta^*}$ uniformly approximates the true function $m = \mu \circ w$ over $\{x: \|x\| \leq 1\}$.

Since $w$ is continuous, by the universal approximation theorem \citep{cybe:89,horn:91}, for any $\delta > 0$, there exists a neural network $w_{\theta^*}$ such that:
\[\sup_{\|x\| \leq 1} \| w_{\theta^*}(x) - w(x) \|_2 < \delta.\]
The weighted \f mean map $\mu:\Delta^{n-1}\mapsto\Omega$ is continuous by Lemma~\ref{lem:frechet-continuity}. Since both $w_{\theta^*}$ and $w$ are continuous, their images over $\{x: \|x\|\leq 1\}$ form a compact subset of $\mathbb{R}^n$. Hence, the continuity of $\mu$ implies uniform continuity on this set. Therefore, for any $\epsilon > 0$, there exists $\delta > 0$ such that
\[
\| w_{\theta^*}(x) - w(x) \|_2 < \delta \quad \Rightarrow \quad d(\mu(w_{\theta^*}(x)), \mu(w(x))) < \epsilon.
\]
We conclude that for any $\epsilon>0$, there exists a neural network $w_{\theta^*}$ such that
\[
\sup_{\|x\| \leq 1} d(m_{\theta^*}(x), m(x)) = \sup_{\|x\| \leq 1} d(\mu(w_{\theta^*}(x)), \mu(w(x))) < \epsilon.
\]

Now consider the stochastic case. Since $X$ is stochastically bounded, for any $\delta > 0$, there exists a constant $M_\delta > 0$ such that  $P(\|X\|\leq M_\delta) > 1 - \delta$. Repeating the same argument over the compact set $\{X: \|X\|\leq M_\delta\}$, we conclude
\[
P\big(d(m_{\theta^*}(X), m(X)) < \epsilon\big) > 1 - \delta.
\]

This completes the proof.
\end{proof}

\subsection{Proof of Lemma~\ref{lem:frechet-lipschitz}}
\begin{proof}
Let $\mu_1 = \mu(w_1)$ and $\mu_2 = \mu(w_2)$ be the weighted \f means corresponding to weights $w_1, w_2 \in \Delta^{n-1}$. We apply the variance inequality for Hadamard spaces \citep[Proposition 4.4]{stur:03}. For $\mu_1 = \mu(w_1)$ and any $z \in \Omega$, we have:
\[
\sum_{i=1}^n w_{1,i} d^2(z, Y_i)-\sum_{i=1}^n w_{1,i} d^2(\mu_1, Y_i) \geq d^2(z, \mu_1).
\]
Taking $z = \mu_2$ yields:
\begin{equation}\label{eq:m2m1}
    \sum_{i=1}^n w_{1,i} d^2(\mu_2, Y_i)-\sum_{i=1}^n w_{1,i} d^2(\mu_1, Y_i) \geq d^2(\mu_1, \mu_2).
\end{equation}

Similarly, for $\mu_2 = \mu(w_2)$ and $z = \mu_1$:
\begin{equation}\label{eq:m1m2}
    \sum_{i=1}^n w_{2,i} d^2(\mu_1, Y_i)-\sum_{i=1}^n w_{2,i} d^2(\mu_2, Y_i) \geq d^2(\mu_1, \mu_2).
\end{equation}

Adding \eqref{eq:m2m1} and \eqref{eq:m1m2}, we obtain:
\begin{equation}\label{eq:mu-upper}
\sum_{i=1}^n (w_{1,i} - w_{2,i}) \left(d^2(\mu_2, Y_i) - d^2(\mu_1, Y_i)\right) \geq 2 d^2(\mu_1, \mu_2).
\end{equation}

Let $a_i = w_{1,i} - w_{2,i}$ and $b_i = d^2(\mu_2, Y_i) - d^2(\mu_1, Y_i)$. Applying the Cauchy-Schwarz inequality:
\[
\left| \sum_{i=1}^n a_i b_i \right| \leq \left( \sum_{i=1}^n a_i^2 \right)^{1/2} \left( \sum_{i=1}^n b_i^2 \right)^{1/2} = \|w_1 - w_2\|_2 \left( \sum_{i=1}^n \left(d^2(\mu_2, Y_i) - d^2(\mu_1, Y_i)\right)^2 \right)^{1/2}.
\]

To bound the $b_i$ terms, observe that by the triangle inequality:
\[
|d^2(\mu_2, Y_i) - d^2(\mu_1, Y_i)| \leq \left|d(\mu_2, Y_i) - d(\mu_1, Y_i)\right|\left(d(\mu_2, Y_i) + d(\mu_1, Y_i)\right) \leq 2Dd(\mu_1, \mu_2),
\]
where $D = \sup_{u,v \in \Omega}d(u,v)$ is the diameter of $\Omega$ and is finite by boundedness.

Therefore,
\[
\left(d^2(\mu_2, Y_i) - d^2(\mu_1, Y_i)\right)^2 \leq 4D^2d^2(\mu_1, \mu_2),
\]
and
\[
\sum_{i=1}^n \left(d^2(\mu_2, Y_i) - d^2(\mu_1, Y_i)\right)^2 \leq 4D^2nd^2(\mu_1, \mu_2).
\]

So we conclude:
\[
\left| \sum_{i=1}^n (w_{1,i} - w_{2,i}) \left(d^2(\mu_2, Y_i) - d^2(\mu_1, Y_i)\right) \right| \leq 2D\sqrt{n} \, \|w_1 - w_2\|_2d(\mu_1, \mu_2).
\]

Combining with \eqref{eq:mu-upper},
\[
2D\sqrt{n} \, \|w_1 - w_2\|_2d(\mu_1, \mu_2) \geq 2 d^2(\mu_1, \mu_2).
\]

Divide both sides by $2 d(\mu_1, \mu_2)$ (noting that the inequality holds trivially if $d(\mu_1, \mu_2) = 0$), we get
\[
d(\mu_1, \mu_2) \leq D\sqrt{n} \, \|w_1 - w_2\|_2,
\]
which completes the proof.
\end{proof}

\subsection{Proof of Theorem~\ref{thm:adam}}
\begin{proof}
We first show that the loss $\ell(\theta; (X, Y))$ is bounded below. By definition,
\[
\ell(\theta; (X, Y)) = d^2(m_\theta(X), Y) + \lambda H(w_\theta(X)),
\]
where $m_\theta = \mu\circ w_\theta$. Since $d^2(\cdot, \cdot) \geq 0$, it suffices to lower bound the second term.

Recall that
\[
H(w) = -\sum_{i=1}^n w_i \log(w_i + \delta),
\]
where $\delta > 0$ is a small regularization parameter to avoid taking the log of zero. The function $H(w)$ is minimized when the distribution $w$ is as concentrated as possible, that is, when $w_i = 1$ for some $i$ and $w_j = 0$ for all $j \neq i$. Therefore, for any $w \in \Delta^{n-1}$,
\[
-\log(1+\delta) \leq H(w) \leq -\log(\delta).
\]
Thus, for all $\theta$,
\[
\ell(\theta; (X, Y)) \geq -|\lambda| \max\{|\log(1+\delta)|, |\log(\delta)|\},
\]
and taking expectations yields
\[
\mathcal{L}(\theta) = E[\ell(\theta; (X, Y))] \geq -|\lambda| \max\{|\log(1+\delta)|, |\log(\delta)|\}.
\]

We now establish the Lipschitz continuity of $\ell(\theta; (X, Y))$ with respect to $\theta$. By Assumption~\ref{ass:generalization}(i), $w_\theta$ is $L$-Lipschitz. %The softmax function $w$ is $1$-Lipschitz \citep{gao:17}.
Moreover, by Lemma~\ref{lem:frechet-lipschitz}, the weighted \f mean map $\mu: \Delta^{n-1} \mapsto \Omega$ is $D\sqrt{n}$-Lipschitz, where $D = \sup_{u,v \in \Omega}d(u,v)$ is the diameter of $\Omega$. Thus, the overall map $m_\theta = \mu\circ w_\theta$ is $LD\sqrt{n}$-Lipschitz continuous with respect to $\theta$.

The squared distance term $d^2(m_\theta(X), Y)$ satisfies
\begin{align*}
    &|d^2(m_{\theta_2}(X), Y) - d^2(m_{\theta_1}(X), Y)| \\&=|d(m_{\theta_2}(X), Y) - d(m_{\theta_1}(X), Y)|(d(m_{\theta_2}(X), Y) + d(m_{\theta_1}(X), Y)),
\end{align*}
and using the triangle inequality,
\[
|d(m_{\theta_2}(X), Y) - d(m_{\theta_1}(X), Y)| \leq d(m_{\theta_2}(X), m_{\theta_1}(X)).
\]
Since $d(m_{\theta_2}(X), Y), d(m_{\theta_1}(X), Y) \leq D$, we obtain
\[
|d^2(m_{\theta_2}(X), Y) - d^2(m_{\theta_1}(X), Y)| \leq 2D d(m_{\theta_2}(X), m_{\theta_1}(X)).
\]
Thus, the contribution of the distance term to the Lipschitz constant is at most $2LD^2\sqrt{n}$.

Next, we bound the entropy term. Recall
\[
H(w) = -\sum_{i=1}^n w_i \log(w_i + \delta),
\]
and its gradient with respect to $w$ has coordinates
\[
\frac{\partial H}{\partial w_i} = -\log(w_i + \delta) - \frac{w_i}{w_i+\delta}.
\]
Since $w_i \in [0,1]$, we have
\[
\left| \frac{\partial H}{\partial w_i} \right| \leq |\log \delta| + 1.
\]
Thus, the gradient of $H$ is bounded in $\ell_\infty$ norm by $|\log \delta| + 1$, and in $\ell_2$ norm by
\[
\|\nabla H(w)\|_2 \leq \sqrt{n} (|\log \delta| + 1).
\]
Therefore,
\[
|H(w_{\theta_2}(X)) - H(w_{\theta_1}(X))| \leq \sqrt{n} (|\log \delta| + 1) L \|\theta_2 - \theta_1\|_2.
\]
Multiplying by $\lambda$ gives a contribution of $|\lambda| \sqrt{n} (|\log \delta| + 1) L$ to the Lipschitz constant.

Summing the contributions of the two terms, the total Lipschitz constant is
\[
G = L\sqrt{n} \left(2D^2 + |\lambda| (|\log \delta| + 1)\right).
\]
Since $\ell$ is differentiable and Lipschitz continuous with constant $G$, we have
\[
\|\nabla_\theta \ell(\theta; (X, Y))\|_2 \leq G.
\]

Finally, applying Corollary~2 from \citet{zahe:18} yields
\[
E\left[\|\nabla \mathcal{L}(\theta_\tau)\|_2^2\right] \leq O\left(\frac{1}{T} + \frac{1}{b}\right),
\]
completing the proof.
\end{proof}

%%%%%%%%%%%%%%%%%%%%%%%%%%%%%%%%%%%%%%%%%%%%%%%%%%%%%%%%%%%%

\section{Choice of Hyperparameters}
\label{app:hyper}
The hyperparameters for E2M can be selected using a grid search over the candidate values listed in Table \ref{tab:hyper}. The optimal combination of hyperparameters is chosen to minimize the mean squared prediction error for the validation data.

\begin{table}[tb]
\centering
\caption{Hyperparameter settings.}
\label{tab:hyper}
\begin{tabular}{lccccc}
\toprule
Regularization parameter & -0.01 & -0.001 & 0 & 0.001 & 0.01 \\
Number of hidden layer & 2 & 3 & 4 & 5 & 6 \\
Number of neurons & 8 & 16 & 32 & 64 & 128 \\
\bottomrule
\end{tabular}
\end{table}

\section{Sensitivity Analysis on Entropy Regularization}
\label{app:entropy}
Entropy regularization controls the sharpness of the learned weight distribution. Negative values of the regularization parameter $\lambda$ encourage higher-entropy (more uniform) weights, leading to a global smoothing effect. Positive values of $\lambda$ favor lower-entropy (more concentrated) weights, pushing the model toward stronger localization.

To assess robustness, we conducted a sensitivity analysis under the same distributional simulation setup as in Section~\ref{sec:sim}, fixing the neural network to two hidden layers with eight neurons each. The entropy regularization parameter was varied over
\[\lambda \in \{-0.1, -0.05, -0.01, 0, 0.01, 0.05, 0.1\},\]
and performance was evaluated in terms of MSPE averaged over 200 Monte Carlo replications. Table~\ref{tab:entropy_sensitivity} reports the average MSPE along with standard deviations.

\begin{table}[tb]
\centering
\caption{Average mean squared prediction errors with standard deviations (in parentheses) for sensitivity analysis of entropy regularization in distributional outputs.}
\label{tab:entropy_sensitivity}
\begin{tabular}{c|ccccccc}
\toprule
$n$ & $-0.1$ & $-0.05$ & $-0.01$ & $0$ & $0.01$ & $0.05$ & $0.1$ \\
\midrule
500  & 0.661 & 0.557 & 0.552 & 0.717 & 0.718 & 0.974 & 1.154 \\
     & (0.172) & (0.174) & (0.195) & (0.199) & (0.163) & (0.122) & (0.177) \\
1000 & 0.507 & 0.405 & 0.368 & 0.509 & 0.584 & 0.886 & 1.104 \\
     & (0.105) & (0.116) & (0.131) & (0.176) & (0.169) & (0.163) & (0.173) \\
2000 & 0.457 & 0.341 & 0.291 & 0.444 & 0.492 & 0.816 & 1.033 \\
     & (0.071) & (0.080) & (0.104) & (0.136) & (0.140) & (0.187) & (0.210) \\
\bottomrule
\end{tabular}
\end{table}

The sensitivity analysis reveals several clear trends. The best performance across all sample sizes occurs at $\lambda = -0.01$, indicating that mild negative regularization provides the most effective balance between global smoothing and local adaptivity. Slightly less negative values such as $\lambda = -0.05$ also perform well, while stronger negative regularization (e.g., $\lambda = -0.1$) leads to oversmoothing and reduced accuracy at larger sample sizes. When $\lambda \geq 0$, performance deteriorates because the model already incorporates natural localization through the softmax weighting, and additional positive entropy penalties force the weights to concentrate further. This over-concentration reduces the effective sample size, increases variance and leads to poor generalization, particularly for large positive values such as $\lambda = 0.05$ or $0.1$.

Overall, these results confirm that \textsc{E2M} is robust to moderate variations in $\lambda$ and benefits most from mild negative values that encourage balanced weighting across training samples.

\section{Additional Experiments for Small Sample Sizes}\label{app:small}
To examine the performance of E2M in small-sample regimes, we conduct additional experiments with sample sizes $n \in \{50, 100, 200\}$ under the same distributional simulation setup as in Section~\ref{sec:sim}. Model architectures and tuning parameters are selected using the same cross-validation procedures as in the main experiments. Performance is evaluated using mean squared prediction error over repeated simulation runs.

The results, summarized in Table~\ref{tab:small}, show that E2M remains competitive for $n=50$ and $n=100$, achieving prediction errors comparable to strong baseline methods such as single-index \f regression (IFR) and global \f regression (GFR). As the sample size increases, the advantage of E2M becomes more pronounced: when $n=200$, E2M attains the lowest mean squared prediction error among all compared methods. These findings indicate that E2M does not rely on very large sample sizes to perform well and can already deliver strong predictive performance at relatively small sample sizes in practice.
\begin{table}[tb]
\centering
\caption{Average mean squared prediction errors (mean on first line, standard deviation in parentheses on second line) of E2M, deep \f regression (DFR) \citep{iao:25}, global \f regression (GFR) \citep{mull:19:6}, sufficient dimension reduction (SDR) \citep{zhan:21:1} and single index \f regression (IFR) \citep{mull:23:3} for distributional outputs in small-sample regimes.}
\label{tab:small}
\begin{tabular}{c|ccccc}
\toprule
$n$ & E2M & DFR & GFR & SDR & IFR \\
\midrule
50  & 1.151 & 1.832 & 1.433 & 4.983 & \textbf{1.103}\\
& (0.154) & (0.411) & (0.254) & (10.645) & (0.229)\\
100  & 1.048 & 1.720 & 1.022 & 1.741 & \textbf{1.002}\\
& (0.135) & (0.311) & (0.112) & (1.315) & (0.254)\\
200  & \textbf{0.851} & 1.634 & 0.855 & 1.135 & 0.971\\
& (0.158) & (0.289) & (0.076) & (0.887) & (0.240)\\
\bottomrule
\end{tabular}
\end{table}

\section{Additional Simulation under Linear Relationships}\label{app:linear}
To further assess the behavior of E2M in settings where the true input-output relationship is linear, we conducted an additional simulation study. This experiment addresses the question of how E2M performs when the data-generating mechanism aligns with classical linear regression.

We consider Gaussian distributional outputs as in Section~\ref{sec:sim}, but modify the conditional mean structure to be linear in the predictors. The input vector $X \in \mathbb{R}^{12}$ is generated from a multivariate normal distribution with mean zero and Toeplitz covariance structure $\Sigma_{ij}=0.8^{|i-j|}$. The conditional mean of the Gaussian distribution is specified as
\[\mu(X) = \sum_{j=1}^{12} \beta_j X_j,\]
with coefficient vector
\[\beta = (1,0,1,0,1,-1,1,-1,1,0,1,0),\]
while the conditional standard deviation is constant, $\sigma(X) = 2$.
To mimic practical settings, we generate $N$ independent samples from each Gaussian distribution and construct empirical distributions as noisy proxies for the latent outputs, following the same procedure as in Section~\ref{sec:sim}.

Table~\ref{tab:linear} reports the mean squared prediction errors across different sample sizes $n \in \{50,100,200,500,1000,2000\}$. In this linear setting, global \f regression (GFR), which directly models a linear relationship between predictors and the conditional \f mean, achieves the lowest prediction error across all sample sizes. Sufficient dimension reduction (SDR) also performs well. E2M remains competitive and improves as $n$ increases, but does not outperform GFR or SDR in this regime. These findings are consistent with the design of E2M. While E2M provides flexible nonlinear modeling capacity, it does not offer a structural advantage in strictly linear settings where simpler models are correctly specified. In such cases, global \f regression is statistically more efficient due to its lower variance and reduced model complexity. This result further highlights that the strength of E2M lies in capturing nonlinear input-output relationships rather than improving performance in settings already well described by linear models.

\begin{table}[tb]
\centering
\caption{Average mean squared prediction errors (mean on first line, standard deviation in parentheses on second line) of E2M, deep \f regression (DFR) \citep{iao:25}, global \f regression (GFR) \citep{mull:19:6}, sufficient dimension reduction (SDR) \citep{zhan:21:1} and single index \f regression (IFR) \citep{mull:23:3} for distributional outputs under linear relationships.}
\label{tab:linear}
\begin{tabular}{c|ccccc}
\toprule
$n$ & E2M & DFR & GFR & SDR & IFR \\
\midrule
50  & 1.388 & 2.167 & \textbf{0.463} & 1.834 & 1.587\\
  & (0.591) & (0.586) & (0.178) & (1.443) & (0.841)\\
100  & 0.814 & 1.409 & \textbf{0.204} & 0.609 & 1.556\\
  & (0.289) & (0.300) & (0.083) & (0.363) & (0.708)\\
200  & 0.589 & 1.173 & \textbf{0.102} & 0.275 & 1.689\\
  & (0.205) & (0.241) & (0.030) & (0.124) & (0.942)\\
500  & 0.391 & 1.041 & \textbf{0.039} & 0.151 & 1.611\\
  & (0.160) & (0.227) & (0.009) & (0.104) & (0.740)\\
1000  & 0.286 & 0.821 & \textbf{0.022} & 0.095 & 1.189\\
 & (0.135) & (0.129) & (0.005) & (0.085) & (0.538)\\
2000  & 0.188 & 0.567 & \textbf{0.015} & 0.074 & 1.492\\
 & (0.086) & (0.082) & (0.004) & (0.061) & (0.605)\\
\bottomrule
\end{tabular}
\end{table}

\section{Permutation-Based Variable Importance}\label{app:vi}
While E2M is primarily designed for flexible prediction in general metric spaces, it is also of interest to assess which covariates contribute most to predictive performance. To this end, we employ a permutation-based variable importance measure, adapted to the \f regression setting. This approach is model-agnostic and has been widely used in machine learning for assessing predictive relevance \citep{fish:19, moln:25}.

Let $\widehat{m}$ denote the trained E2M predictor and let $\mathcal{D}_{\text{test}} = \{(X_i, Y_i)\}_{i=1}^{n_{\text{test}}}$ denote a test dataset. The baseline prediction error is defined as
\[
\mathrm{Err}_{\text{base}}
=
\frac{1}{n_{\text{test}}}
\sum_{i=1}^{n_{\text{test}}}
d^2\bigl(Y_i, \widehat{m}(X_i)\bigr),
\]
where $d$ denotes the metric on the output space.

For each covariate, we generate $B$ independent random permutations. For permutation $b \in \{1,\dots,B\}$, let $X_i^{(j,b)}$ denote the predictor vector in which the $j$-th coordinate has been randomly permuted across test samples, while all other coordinates remain unchanged. The corresponding prediction error is
\[
\mathrm{Err}_{j}^{(b)}
=
\frac{1}{n_{\text{test}}}
\sum_{i=1}^{n_{\text{test}}}
d^2\bigl(Y_i, \widehat{m}(X_i^{(j,b)})\bigr).
\]

The permutation importance score for covariate $X_j$ is defined as the average increase in prediction error:
\[
\mathrm{VI}_j
=
\frac{1}{B}
\sum_{b=1}^{B}
\mathrm{Err}_{j}^{(b)} - \mathrm{Err}_{\text{base}}.
\]

A larger value of $\mathrm{VI}_j$ indicates that permuting $X_j$ leads to a greater degradation in predictive performance, suggesting higher predictive importance.

\vskip 0.2in
\bibliography{collection}

\end{document}